\documentclass{article}

\usepackage{microtype}
\usepackage{graphicx}
\usepackage{subcaption}
\usepackage{caption}
\usepackage{booktabs} 
\usepackage{multirow}

\usepackage{hyperref}

\usepackage[accepted]{icml2025}

\usepackage{amsmath}
\usepackage{amssymb}
\usepackage{mathtools}
\usepackage{amsthm}
\usepackage{amsfonts}

\usepackage[capitalize,noabbrev]{cleveref}

\usepackage{makecell}
\usepackage{wrapfig}
\usepackage{float}

\theoremstyle{plain}
\newtheorem{theorem}{Theorem}[section]

\theoremstyle{definition}
\newtheorem{definition}[theorem]{Definition}

\theoremstyle{remark}

\usepackage{booktabs}
\usepackage{colortbl}
\PassOptionsToPackage{table,xcdraw}{xcolor}
\usepackage{xcolor}

\newcommand{\name}{\textsc{Locret}}
\newcommand{\nameq}{\textsc{Locret-Q}}
\newcommand{\rev}{\textcolor{black}}

\usepackage[textsize=tiny]{todonotes}

\icmltitlerunning{\name: Enhancing Eviction in Long-Context LLM Inference with Trained Retaining Heads on Consumer-Grade Devices}

\begin{document}

\twocolumn[
\icmltitle{\name: Enhancing Eviction in \underline{Lo}ng-\underline{C}ontext LLM Inference\\ with Trained \underline{Ret}aining Heads on Consumer-Grade Devices}




\begin{icmlauthorlist}
\icmlauthor{Yuxiang Huang}{thu}
\icmlauthor{Binhang Yuan}{ust}
\icmlauthor{Xu Han}{thu}
\icmlauthor{Chaojun Xiao}{thu}
\icmlauthor{Zhiyuan Liu}{thu}
\end{icmlauthorlist}

\icmlaffiliation{thu}{Department of Computer Science and Technology, Institute for Artificial Intelligence, Beijing National Research Center for Information Science and Technology, Tsinghua University, Beijing, China.}
\icmlaffiliation{ust}{Department of Computer Science and Engineering, The Hong Kong University of Science and Technology, Hong Kong, China.}

\icmlcorrespondingauthor{Binhang Yuan}{biyuan@ust.hk}
\icmlcorrespondingauthor{Xu Han}{han-xu@tsinghua.edu.cn}

\icmlkeywords{Long-context Inference, Memory Efficient Inference, Large Language Models}

\vskip 0.3in
]



\printAffiliationsAndNotice{}  

\begin{abstract}
Scaling the input context length of a large language model (LLM) incurs a significant increase in computation cost and memory footprint to maintain the attention key-value (KV) cache.
Existing KV cache compression methods suffer from inefficient compression strategies and limited memory reduction effects, making it difficult for LLMs to conduct long-context inference on consumer-grade devices, especially when inferring long-context stream input.
Such obstacles prevent consumer-grade devices from supporting more complex applications, creating challenges for the democratization of LLMs.
To overcome this, we propose \name, the first framework to create an eviction policy compatible with chunked prefill. By evaluating the causal importance of KV cache units by learnable \textit{retaining heads}, \name~enables precise eviction of cache units, facilitating efficient long-context inference.
In our extensive empirical studies, \name~outperforms the recent popular and competitive approaches in terms of memory efficiency and generation quality
--- \name~achieves up to $20\times$ of KV cache compression ratio within less than $10\%$ performance loss.
Furthermore, \name~achieves 128K+ long-context inference on a single NVIDIA 4090 GPU without compromising generation quality and only costs $<1$ GPU hour of additional training.
We provide the code of \name~at \url{https://github.com/huangyuxiang03/Locret}. 
\end{abstract}

\section{Introduction}


In recent years, large language models (LLMs) have revolutionized generative AI~\citep{zhao2023survey,minaee2024large}, and the advancements of LLMs in handling long-context tasks have further unlocked the potential of generative AI. As a result, the context lengths supported by state-of-the-art LLMs have been significantly extended, such as GPT-4o~\citep{gpt4o} handling 128K tokens, Claude-3~\citep{claude3} supporting 200K tokens, and Gemini-1.5~\citep{reid2024gemini} even reaching 10M tokens. These improvements enable LLMs to tackle complex applications with extremely long or streaming inputs, such as multi-hop reasoning~\citep{li2024making,schnitzler2024morehopqa}, LLM-driven agents~\citep{qin2023tool,wang2024survey}, and AI-powered operating systems~\citep{mei2024aios}. 
Some recent efforts~\cite{hu2024minicpm,abdin2024phi} have successfully deployed LLMs on consumer-grade end-side devices instead of cloud servers and conducted inference with limited context.
We envision that unleashing the potential of long-context inference on consumer-grade devices will revolutionize the development of personalized AI applications and the democratization of LLMs.
However, \textit{conducting long-context LLM inference on consumer-grade devices remains a challenging problem that requires algorithmic innovations and systematic optimizations.}

\begin{figure}[t]
  \centering
  \includegraphics[width=0.65\linewidth]{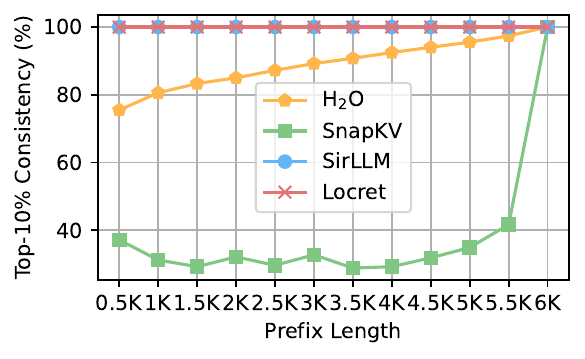} 
\vspace{-10pt}
  \caption{
  For each prefix length of the context, this figure shows the \textit{consistency} in evaluating the token importance of the prefix based on the full context and based on only the prefix without subsequent tokens. The \textit{consistency} is defined as the intersection of the top 10\% tokens of two evaluation methods divided by the number of top 10\% tokens in the prefix.
  More details are in Appendix~\ref{append:discrepancy}.
  }
    \label{fig:consistency}
\vspace{-15pt}
\end{figure}

As context length scales, the challenge of long-context LLM inference arises from two major aspects: the increased computational cost of the attention mechanism and the higher memory footprint due to the key-value (KV) cache.
This leads to the failure of traditional optimizations targeting model backbones to provide a sufficient solution. 
Specifically, backbone-targeted optimizations, such as compacting model architectures~\citep{hu2024minicpm,abdin2024phi} and quantizing model weights~\citep{frantar2022gptq,dettmers2022gpt3,xiao2023smoothquant,lin2024awq}, fail to improve the efficiency of attention patterns or the KV cache, as attention's quadratic complexity with respect to sequence length remains unaddressed.
To this end, recent efforts have focused on optimizing attention patterns and the KV cache to achieve efficient long-context LLM inference.


Recent attention-targeting optimizations, including sparse attention~\citep{jiang2024minference, ge2023model, lou2024sparser} and KV cache quantization~\citep{liu2024kivi, hooper2024kvquant, zandieh2024qjl}, show promising results to accelerate attention computation and reduce memory footprint. 
However, they fail to fundamentally address the core challenge: \textit{the KV cache grows linearly with context length}. 
Layer-wise chunked prefill combined with attention sparsity (e.g., \textsc{SnapKV}~\citep{li2024snapkv}) can alleviate this problem to a certain extent.
This technique typically performs cache eviction after the precise attention computation for each layer and requires access to the entire sequence.
It can theoretically support longer sequences by limiting the maximum memory usage to a single layer's KV cache, but \textit{it cannot handle streaming input whose length grows continually}. 
The combination of cache eviction methods~\citep{xiao2023efficient,yang2024pyramidinfer} and chunked prefill offers a more promising approach by maintaining a static cache size and supporting streaming input. 
Yet, as shown in Figure~\ref{fig:consistency}, existing eviction techniques like \textsc{H$_2$O}~\citep{zhang2024h2o} and \textsc{SnapKV} show significant discrepancies between local importance estimation and global importance estimation, i.e. it is hard to estimate the importance of each token only based its previous tokens. Instead, these methods require a large number of subsequent tokens to make an accurate estimation.
Other methods like \textsc{SirLLM}~\citep{yao2024sirllm} show great local-global estimation consistency but suffer from performance degradation.

To overcome these limitations, we propose a lightweight training-based paradigm, \name, that provides a more accurate token importance estimation to select the victim during KV cache eviction, enabling efficient and scalable long-context LLM inference:

\underline{\textbf{Contribution 1:}} 
We propose \name, a lightweight training-based paradigm for selective KV cache eviction in long-context LLM inference. 
It introduces learnable \textit{retaining heads} to estimate the \textit{causal importance score} (CIS) for token selection, with an offline training cost of $<$1 GPU hour.
Additionally, we present \nameq, a query-aware variant of \name, slightly modified to handle query-driven tasks (e.g., long document question answering).


\underline{\textbf{Contribution 2:}} 
We provide an efficient inference system implementation for \name, integrating retaining heads into a chunked prefill framework. 
This integration limits GPU memory usage by evicting low-CIS KV cache units during the prefill process and accelerates the prefill time. 
\name~is compatible with all decoder-only LLMs and imposes minimal additional hardware requirements.


\underline{\textbf{Contribution 3:}} 
We extensively evaluate \name, demonstrating its ability to achieve comparable performance with full KV cache while maintaining inference efficiency. \name~achieves over $20\times$ and $8\times$ KV cache compression ratios for \texttt{Phi-3-mini-128K} and \texttt{Llama-3.1-} \texttt{8B-instruct}, respectively. 
Additionally, \nameq~accelerates prefill by over $2\times$ on query-driven tasks without significant performance degradation. 
This framework enables full comprehension of long contexts on consumer-grade devices without compromising the generation quality, and introduces minimal additional system optimizations.

\section{Related Work}
\label{sec:related_work}
\label{sec:related_work_compression}

This paper focuses on optimizing long-context LLM inference.
Existing efforts can be categorized into algorithm and system optimization. 
For more details about LLMs, please refer to the surveys~\cite{zhao2023survey, lu2024small}.

\textbf{Algorithm Optimizations} aim to reduce the size of the KV cache and can generally be classified into three categories: {quantization-based methods}, {sparsity-based methods}, and {token dropping methods}. 
Quantization-based methods~\citep{liu2024kivi,hooper2024kvquant, zandieh2024qjl,zhang2024kv} use low-bit values to represent the KV cache, reducing cache memory overhead and improving cache computing efficiency. These quantization-based methods suffer from hardware-oriented operator customization and additional inverse quantization overhead.
Sparsity-based methods~\citep{ge2023model,jiang2024minference, yang2024pyramidinfer,lou2024sparser,lv2024critiprefill} leverage the sparsity patterns of attention heads to reduce both computational and I/O costs. Combining different patterns can yield further optimization by identifying specific patterns for each head~\cite{ge2023model, jiang2024minference, xiao2024duoattention}. For more details on sparsity-based methods, please refer to the surveys~\cite{yuan2024kv,kang2024gear,luohe2024keep}. Although quantization-based methods and sparsity-based methods have achieved promising results, they cannot address the issue that the KV cache memory overhead increases linearly with the context length. Eviction-based methods, such as \textsc{H$_2$O}~\citep{zhang2024h2o}, \textsc{ScissorHands}~\citep{liu2024scissorhands}, and \textsc{SirLLM}~\citep{yao2024sirllm}, rank KV cache units by certain statistical metrics to identify the most influential units, discarding others to reduce memory usage and speed up attention computation. Pooling-based methods~\citep{nawrot2024dynamic,rajput2024infer}, especially \textsc{StreamingLLM}~\citep{xiao2023efficient} and \textsc{LoCoCo}~\citep{cai2024lococo}, compress multiple adjacent KV cache units into a single unit using pre-designed transformations. More important units will merge into compressed units with higher weights.
Eviction-based and pooling-based methods drop or merge tokens to maintain a static cache size, but struggle with accurate victim selection and optimal pooling function design.

\textbf{System Optimizations} alleviate the challenge of long-context inference from a system-level perspective, by fully considering hardware features. Offloading-based methods~\citep{sheng2023flexgen,xiao2024infllm,wu2024offload,sun2024shadowkv} use CPU memory to store the KV cache and retrieve only the most relevant chunks to GPU memory before inferring a new chunk. These methods reduce maximum GPU memory usage at the cost of introducing CPU-GPU communication overhead. 
Hardware-aware methods, such as \textsc{Flash-Atttention}~\citep{dao2022flashattention,dao2023flashattention2,shah2024flashattention} and \textsc{Page-Attention}~\citep{kwon2023efficient}, enable more efficient runtime memory management by considering GPU architectures~\citep{ghorpade2012gpgpu}.
In addition, building inference infrastructures with a more efficient programming language~\citep{llamacpp,llama2c,rustformers}, or adopting disaggregated inference~\citep{jianghexgen,zhong2024distserve,qin2024mooncake,hu2024inference}, can also greatly improve long-context inference efficiency. 
Since system optimizations primarily enhance efficiency by leveraging hardware resources rather than directly optimizing attention patterns or the KV cache, relying solely on them cannot adequately address the challenges of long-context LLM inference. 
Several efforts have integrated algorithm optimizations into system optimizations~\citep{agrawal2023sarathi,lee2024infinigen}, such as \textsc{KTransformers}~\citep{ktransformers} leveraging offloading based on \textsc{InfLLM}~\citep{xiao2024infllm}, achieving promising results.

\begin{figure*}[!ht]
\begin{center}
\includegraphics[width=1\linewidth]{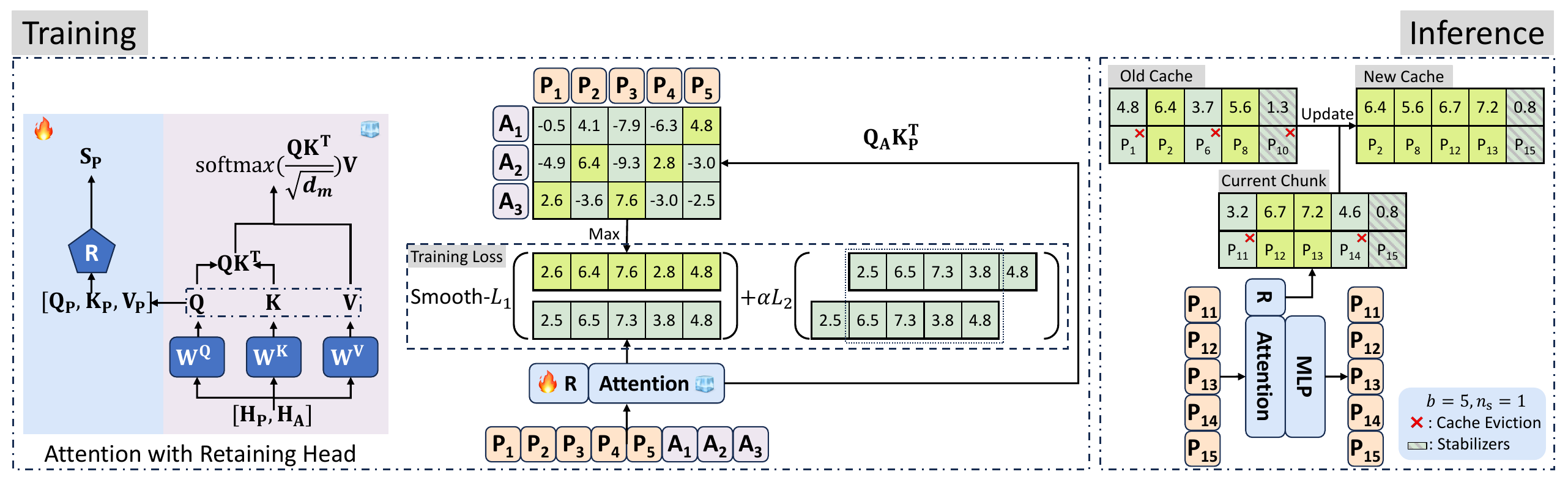}
\end{center}
\vspace{-10pt}
\caption{The framework of \name. ``$\mathbf{R}$'' represents the retaining head. $\text{P}_i$ and $\text{A}_i$ correspond to the $i$-th prompt token and answer token. ``t'' represents the time step in chunked prefill, ``$b$'' represents the budget size, and ``$n_{s}$'' represents the length of the stabilizers. For simplicity, our notation here does not reflect the concept of layers.}
\label{fig:framework}
\vspace{-10pt}
\end{figure*}

\section{Methodology of \name~}

\subsection{Preliminaries}

\textbf{Transformer Architecture.} 
Given a token sequence $\{t_1, \cdots, t_n\}$ as the input prompt of the transformer-based LLM,
we denote the output hidden states of the $i$-th layer as $\textbf{H}^{(i)}$, and denote the input embeddings of the first layer as $\textbf{H}^{(0)}$.
For each transformer layer, it consists of an attention block and a feedforward neural network (FFN) block. Attention blocks often follow the grouped-query attention (GQA) architecture~\citep{ainslie2023gqa}, with $h$ query heads and $h/g$ KV heads, where $g$ is the group size, i.e., $g$ heads share the same KV heads. The multi-head attention (MHA) architecture adopted in the original transformer can be regarded as a special GQA ($g=1$). 
In the $i$-th layer, the attention score of the $j$-th query head is formalized as \scalebox{0.88}{$\mathbf{A}_j^{(i)} = \texttt{softmax}\left(\mathbf{Q}_j^{(i)}{\mathbf{K}_{\lceil {j/g}\rceil}^{(i)\top}}/\sqrt{d_m}\right)\cdot \mathbf{V}_{\lceil{j/g}\rceil}^{(i)}$},
where $d_m$ represents the hidden size for each head and 
\scalebox{0.88}{$\left[\mathbf{Q}_j^{(i)}, \mathbf{K}_{\lceil{j/g}\rceil}^{(i)}, \mathbf{V}_{\lceil{j/g}\rceil}^{(i)}\right] = \mathbf{H}^{(i-1)}\cdot\left[{\mathbf{W}_j^{(i)\texttt{Q}}}, {\mathbf{W}_{\lceil{j/g}\rceil}^{(i)\texttt{K}}}, {\mathbf{W}_{\lceil{j/g}\rceil}^{(i)\texttt{V}}}\right]$}.
After obtaining the attention score, the output of the $i$-th attention block is \scalebox{0.88}{$\mathbf{A}^{(i)} = \left[\mathbf{A}_1^{(i)}, \cdots, \mathbf{A}_h^{(i)}\right]\cdot\mathbf{W}^{(i)\texttt{O}}$}, and the output hidden states of the $i$-th layer is \scalebox{0.88}{$\mathbf{H}^{(i)} = \texttt{FFN}(\mathbf{A}^{(i)})$}.

\textbf{KV Cache and Chunked Prefill.} Given the input prompt sequence $\{t_1, \cdots, t_n\}$, during the prefill stage, all prompt tokens are processed in a single forward pass. After the prefill, \scalebox{0.88}{$\mathbf{K}^{(i)} = \left[\mathbf{K}^{(i)}_1, \cdots, \mathbf{K}^{(i)}_{h/g}\right]$} and \scalebox{0.88}{$\mathbf{Q}^{(i)} = \left[\mathbf{Q}^{(i)}_1, \cdots, \mathbf{Q}^{(i)}_{h/g}\right]$} are stored as the KV cache, whose sequence length is $n$.
During the decoding stage, each time a token is decoded, a forward pass is conducted only for this token and decode the next token. In this process, the KV cache is used to avoid redundant attention computation.
Chunked prefill is a method for reducing peak memory usage by spliting tokens into mulitple chunks and prefilling tokens chunk by chunk.
Taking both the KV cache and chunked prefill into account, the attention block can be modified as follows,
\vspace{-8pt}
\begin{equation}
\small
    \label{eq:attn_decode}
    \begin{aligned}
        \mathbf{A}{[n+1\colon n+B]}_j^{(i)} = \texttt{softmax}\Biggl( & \\ \frac{\mathbf{Q}{[n+1\colon n+B]}_j^{(i)}  \mathbf{K}{[1\colon n+B]}_{\lceil {j/g}\rceil}^{(i)\top}}{{\sqrt{d_m}}} \Biggr) & \mathbf{V}{[1\colon n+B]}_{\lceil{j/g}\rceil}^{(i)},
    \end{aligned}
\end{equation}
where \scalebox{0.88}{$\mathbf{A}[n+1\colon n+B]$} denotes the attention output for the tokens {$\{n+1, \cdots, n+B\}$}, and $B$ is the number of tokens processed in a single forward pass. For decoding, {$B=1$}, while for chunked prefill, $B$ corresponds to the chunk size. 
For the $k$-th token in the context, its attention output is {$\mathbf{A}[k]$}, its key and value vectors are {$\mathbf{K}[k]$} and {$\mathbf{V}[k]$}.

\textbf{Cache Eviction.}
In the cache eviction process, we treat the KV vector pair of a single token within one attention head as the smallest cache unit.
We denote the cache unit of the $k$-th token as $c_k=(\mathbf{K}[k], \mathbf{V}[k])$. 
Assuming a memory budget $b$, representing the maximum number of cache units that can be stored, the abstract form of the attention block can be written as $c_k = f(c_1, c_2, \cdots, c_{k-1})$. 
With limited cache capacity, this process can only be approximated by $\tilde c_k = f(\tilde c_{p_1}, \tilde c_{p_2}, \cdots, \tilde c_{p_{b'}})$, where $b' \leq b$, and $p_1, p_2 \cdots, p_{b'} \in \{1, 2, \cdots, k-1\}$. When the cache is full, one unit must be evicted. We select the victim using some policy $p_v = \text{Policy}(\tilde c_{p_1}, \cdots, \tilde c_{p_b}; \tilde c_k)$,
and the key challenge is to develop an effective policy that minimizes the error $\| \tilde c_k - c_k \|$.

\begin{figure*}[!ht]
\begin{center}
\subfloat[]{\label{fig:stabilizer_a}\includegraphics[width=0.27\linewidth]{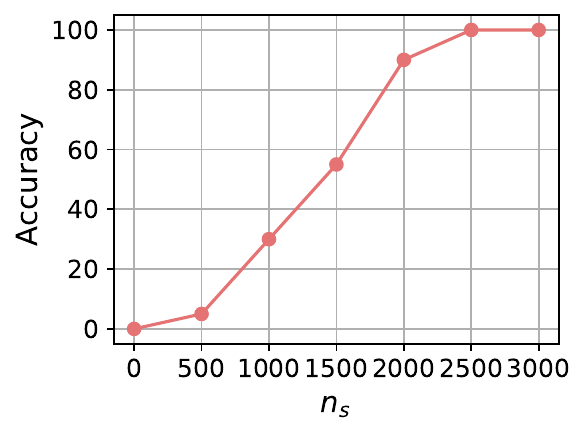}\vspace{-7pt}}
\subfloat[]{\label{fig:stabilizer_b}\includegraphics[width=0.27\linewidth]{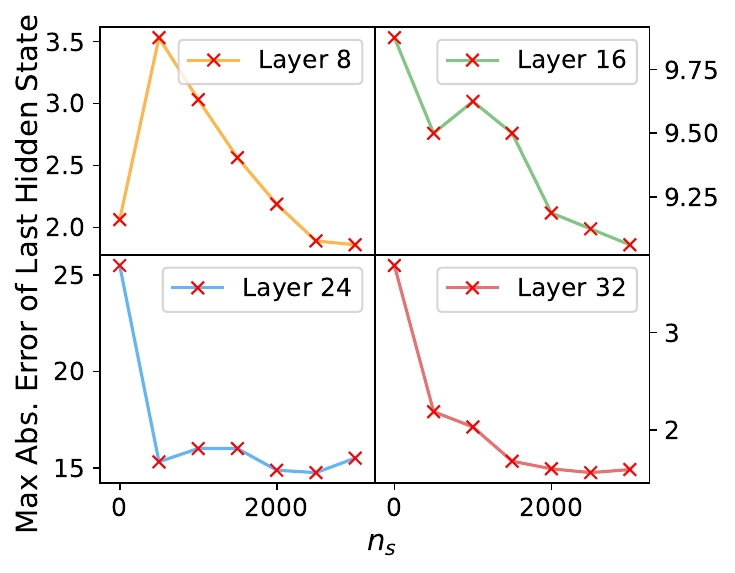}\vspace{-7pt}}
\subfloat[]{\label{fig:stabilizer_c}\includegraphics[width=0.27\linewidth]{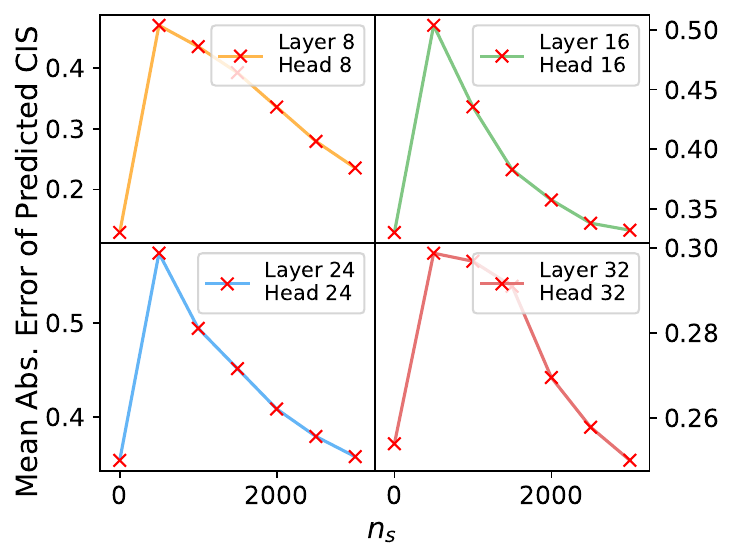}\vspace{-7pt}}
\vspace{-10pt}
\end{center}
\caption{R.Number with different stabilizer lengths $n_s$. (a) Task accuracy under different $n_s$. (b) Maximum absolute error of the last hidden state. (c) Mean absolute error of the predicted CIS. We conduct this experiment on entries 101-120 of R.Number using the \texttt{Phi-3-mini-128K} backbone.}
\label{fig:stabilizer}
\vspace{-15pt}
\end{figure*}

\subsection{Framework of \name}

\name~is a training-based KV cache eviction framework that works in conjunction with chunked prefill. 
As illustrated in Figure~\ref{fig:framework}, \name~operates in two stages: training and inference. 
In the training stage, we modify the original LLM by appending a retaining head $\mathbf{R}$ to each attention module. 
We then train the retaining heads $\mathbf{R}$ while keeping the LLM backbone frozen. 
During the chunked prefill inference stage, the retaining heads $\mathbf{R}$ can obtain the importance of each cache unit. 
We retain the cache units with higher scores, along with stabilizers (i.e., the last tokens), in the cache pool located in GPU memory. 
Through this process, the retaining heads $\mathbf{R}$ learn and predict the patterns discovered by existing methods, e.g. attention sink in~\citet{xiao2023efficient} and vertical lines in \textsc{MInference}~\citep{jiang2024minference}, as detailed in Appendix~\ref{sec:retained_patterns}.

The eviction policy assigns each cache unit an importance score reflecting its influence on comprehending subsequent context. 
This estimation is causal, termed the \textit{causal importance score (CIS)}. The CIS of the $k$-th unit depends only on the preceding units and the $k$-th unit itself. 
Due to memory constraints, calculating the exact CIS on-chip is impractical. 
Intuitively, discarding certain KV cache during the attention calculation introduces errors, manifesting as biases in the CIS. However, as long as the importance estimation is sufficiently accurate, the core information can be preserved, and the loss of information in the attention process can be minimized.
Therefore, the error introduced in the CIS can be neglected.
Please refer to Appendix \ref{append:causal_importance} for more details.



\subsection{Training Retaining Heads}

In this section, we introduce \name's model architecture modifications and the corresponding training recipe. 
We add additional parameters to compute the CIS $\mathbf{S}[k]$ for the $k$-th cache unit.
Specifically, we inject a retaining head, consisting of a small MLP, into each layer. 
From our observation, such small MLPs do not slow down model inference, with details elaborated in Appendix~\ref{append:time}.
The retaining head predicts the CIS for each head of the corresponding layer based on the concatenation of {$[\mathbf{Q}, \mathbf{K}, \mathbf{V}]$}. 
Formally, with a slight abuse of notation, let the retaining head for layer $i$ be denoted as $\mathbf{R}$. The CIS at head $j$ is then calculated as:
\scalebox{0.88}{$\mathbf{\tilde S} = \mathbf{R}([\mathbf{Q}, \mathbf{K}, \mathbf{V}]) = \sigma([\mathbf{Q}, \mathbf{K}, \mathbf{V}]\mathbf{W}_1)\mathbf{W}_2$}.
In this equation, \scalebox{0.88}{$\mathbf{W}_1 \in \mathbb{R}^{(d_m + 2d_{kv}) \times d_{\mathbf{R}}}$} and \scalebox{0.88}{$\mathbf{W}_2 \in \mathbb{R}^{d_{\mathbf{R}} \times \frac{h}{g}}$} are the tunable parameters of $\mathbf{R}$, and $\sigma$ is the activation function, 
\scalebox{0.88}{$\mathbf{\tilde S}[k] = \left[\mathbf{\tilde S}[k]_1, \cdots, \mathbf{\tilde S}[k]_{h/g}\right]\in \mathbb{R}^{\frac{h}{g}}$}, where
\scalebox{0.88}{$\mathbf{\tilde S}[k]_j$} is the predicted CIS of the $k$-th token. 
This architecture implies that the importance estimation for a single head is not performed in isolation but considers all heads.

We train the retaining head $\mathbf{R}$s on a small Question-Answer (QA) supervised fine-tuning (SFT) dataset, where each entry consists of a single prompt and one answer.
We define the CIS $\mathbf{S}[k]_j$ for the $k$-th token at head $j$ as the maximum attention score, before softmax, from all the answer tokens toward the $k$-th token. 
Formally, given a training instance $d$, for the $k$-th token at head $j$ of layer $i$, we approximate the predicted value {$\mathbf{\tilde S}[k]_j^{(i)}$} to the ground truth \scalebox{0.88}{$\mathbf{S}[k]_j^{(i)} := \max_p\left(\mathbf{Q}_j^{(i)}\mathbf{K}_{j}^{(i)T}\right)_{p, k}$}, where {$n_q(d) < p \leq n_q(d) + n_a(d)$}, and {$n_q(d)$} and {$n_a(d)$} represent the lengths of the prompt and answer in data $d$, respectively. For an MHA model with $L$ layers and $h$ heads, the training objective is described in Equation~\ref{eq:obj}. 
For GQA models, we take the maximum attention score before softmax across different query heads within the same group as the ground truth for the corresponding KV head.

\vspace{-18pt}
{\small
\begin{align}
    \label{eq:obj}
    \underset{\mathbf{W}_1^{(i)}, \mathbf{W}_2^{(i)}, i=1, 2\cdots, L}{\text{argmin}} \mathbb{E}_{d\in \mathcal{D}}\left[\sum_{i=1}^{L}\sum_{j=1}^{h}\sum_{k=1}^{n_q(d)}\mathcal{L}\left(\mathbf{\tilde S}[k]_j^{(i)}, 
    \mathbf{S}[k]_j^{(i)}
    \right)\right]
\end{align}
}

\vspace{-13.5pt}

The training loss consists of a regression loss and a smoothing loss. 
We apply the Smooth-$\mathcal{L}_1$ norm between the predicted values and the ground truth. 
Since important segments in natural language typically consist of adjacent tokens, we also apply the $\mathcal{L}_2$ norm between each pair of adjacent predicted values to enforce smoothness. 
The complete training loss for \name~is given by Equation~\ref{eq:loss}.

\vspace{-12.5pt}
{\small
\begin{equation}
\begin{aligned}
    \label{eq:loss}
    \mathcal{L}\left(\mathbf{\tilde S}[k]_j^{(i)}, \mathbf{S}[k]_j^{(i)}\right) &= \text{Smooth-}\mathcal{L}_1\left(\mathbf{\tilde S}[k]_j^{(i)}, \mathbf{S}[k]_j^{(i)}\right) \\&+ \alpha \mathcal{L}_2\left(\mathbf{\tilde S}[k]_j^{(i)}, \mathbf{\tilde S}[k+1]_j^{(i)}\right)
\end{aligned}
\end{equation}
}
\vspace{-12.5pt}

From our observations, the training of \name~exhibits strong robustness. 
Despite changes in $d_{\mathbf{R}}$ and the dataset, the performance variations shown in Figure~\ref{fig:intermediate} and Table~\ref{tab:data-insensitivity} are minimal. 
Details can be found in Appendix~\ref{sec:robust}.
Training statistics and loss dynamics are recorded in Appendix~\ref{sec:training_dynamics}.

\vspace{-5pt}

\subsection{Inference Implementation of \name}

During the inference stage, we use the chunked prefill pattern and perform cache eviction based on the predicted CIS. 
Since the predicted CIS does not rely on subsequent tokens, it remains consistent once calculated.
Thus, we store the KV cache units along with their corresponding causal importance values. 
When the cache is full, we evict the units with lower causal importance values, as they are deemed less useful for future computations. 
Such eviction is performed during chunked prefill. 
When processing a new chunk, we first compute its KV cache, concatenate it with the previously retained cache, and evict redundant units to adhere to the budget size.
Note that we cache the pre-RoPE KV cache and reassign continuous position embeddings from the beginning to enhance context continuity.

Cache eviction introduces context discontinuity, meaning some cache units at certain positions may be absent, which can degrade generation quality.
To mitigate this, we retain the last $n_s$ tokens of the current chunk, named as the \textit{stabilizers}, at each step of chunked prefill, ensuring a local and continuous context to minimize errors. 
As shown in Figure \ref{fig:stabilizer}, smaller $n_s$ results in severe performance degradation, and the model fails entirely when stabilizers are absent, as context discontinuity leads to instability in CIS prediction, causing errors in cache eviction and amplifying errors in hidden states. 
More details are discussed in Appendix \ref{sec:discontinuous-context}. We provide a pseudocode of \name~inference in Algorithm~\ref{alg:infer}.



\section{Experiments}

We conduct experiments to evaluate whether \name~can address the following questions: \:(\underline{\textbf{Q1}}) Can \name~obtain better end-to-end task performance compared to popular and competitive long-context inference methods using similar or less peak memory? \:(\underline{\textbf{Q2}}) Is \name~able to achieve a faster inference speed on consumer-grade devices? \:(\underline{\textbf{Q3}}) How can \name~process query-driven tasks?


\begin{table*}[t]
\small
\caption{The experimental results of \name~compared with all the baselines on $\infty$Bench and L-Eval, where higher score represents better performance. ``Avg.'' represents the average score across all tasks. The highest score in each column is marked in \textbf{bold}, and the second highest is \underline{underlined}. \name~achieves the highest overall score among all competitors.}
\vspace{5pt}
\label{tab:results}
\begin{subtable}{1\textwidth}
\centering
\scalebox{0.84}{
\begin{tabular}{l|ccccccccc|c}
\toprule
Method & R.PassKey & R.Number & E.Sum & E.QA & E.MC & Z.QA & E.Dia & C.Debug & M.Find & Avg.$\uparrow$\\
\midrule
\multicolumn{11}{c}{\texttt{Phi-3-mini-128K} on $\infty$Bench} \\ \midrule
\textsc{FullAttn} & 98.64 & 97.12 & 17.92 & 11.16 & 55.46 & 14.83 & 8.00 & 23.10 & 17.43 & 38.18\\ \midrule
{\textsc{InfLLM}}
& \textbf{100.00} & \underline{97.12} & 14.35 & 4.97 & 38.86 & 11.04 & 3.50 & \underline{25.38} & 15.14 & \underline{34.48}\\ 
\textsc{HF-2bits} & 0.00 & 0.00 & 13.80 & 1.44 & 1.75 & 0.20 & 0.50 & 0.00 & 0.57 & 2.03 \\
\textsc{SirLLM} & 3.39 & 3.39 & \textbf{21.06} & 6.32 & \underline{44.98} & \textbf{11.99} & 5.00 & 22.34 & \underline{21.71} & 15.58\\
\textsc{MInference} & \underline{99.32} & 95.93 & 14.44 & \textbf{8.11} & 40.61 & 10.60 & \underline{9.00} & 15.48 & 15.43 & 32.25\\
\rowcolor{cyan!10}
\textbf{\name~} & \textbf{100.00} & \textbf{97.46} & \underline{16.82} & \underline{7.61} & \textbf{46.29} & \underline{11.31} & \textbf{10.00} & \textbf{27.92} & \textbf{29.71} & \textbf{34.73}\\ \midrule
\multicolumn{11}{c}{\texttt{Llama-3.1-8B-instruct} on $\infty$Bench} \\
\midrule

\textsc{FullAttn} & 100.00 & 99.32 & 26.79 & 15.06 & 68.12 & 13.40 & 17.00 & 20.56 & 34.00 & 43.81\\ \midrule
\textsc{InfLLM} & \textbf{100.00} & \textbf{100.00} & 24.24 & 14.21 & 51.97 & 10.76 & 11.00 & \underline{26.25} & \textbf{35.71} & 41.57\\ 
\textsc{HF-2bits} & 36.78 & 6.95 & 8.77 & 4.05 & 27.95 & 3.09 & 5.50 & 13.20 & 22.00 & 14.25 \\
\textsc{SirLLM} & 1.69 & 1.69 & \underline{25.60} & 8.95 & 55.46 & 10.38 & 9.50 & 23.10 & 3.71 & 15.56\\
{\textsc{MInference}} & \textbf{100.00} & 98.47 & 20.64 & \underline{14.35} & \textbf{59.83} & \textbf{12.20} & \textbf{20.50} & 25.89 & \underline{35.43} & \underline{43.03}\\
\rowcolor{cyan!10}
\textbf{\name~} & \textbf{100.00} & \underline{99.49} & \textbf{27.28} & \textbf{20.90} & \underline{58.82} & \underline{11.85} & \underline{13.00} & \textbf{27.16} & 32.86 & \textbf{43.48}\\
\bottomrule
\end{tabular}
}
\end{subtable}

\begin{subtable}{1\textwidth}
\centering
\scalebox{0.84}{
\begin{tabular}{l|cccccc|c}
\multicolumn{8}{c}{}\\
\toprule
Method & CodeU & NQ & CUAD & NarrativeQA & QMSum & SPACE & Avg.$\uparrow$\\
\midrule
\multicolumn{8}{c}{\texttt{Phi-3-mini-128K} on L-Eval} \\
\midrule

\textsc{FullAttn} & 8.89 & 59.14 & 30.34 & 17.59 & 16.05 & 14.51 & 24.42\\ \midrule
\textsc{InfLLM} & 5.56 & 34.32 & 14.53 & 14.80 & 13.31 & \underline{14.81} & 16.22\\ 
\textsc{HF-2bits} & 0.00 & 1.69 & 6.40 & 2.04 & 2.73 & 3.34 & 2.70\\
{\textsc{SirLLM}} & \textbf{8.89} & \underline{37.92} & 20.89 & 14.51 & 13.70 & 14.46 & \underline{18.40} \\
\textsc{MInference} & \underline{7.78} & 25.21 & \textbf{26.64} & \underline{15.14} & \textbf{15.78} & \textbf{14.87} & 17.57\\
\rowcolor{cyan!10}
\textbf{\name~} & \textbf{8.89} & \textbf{51.49} & \underline{22.23} & \textbf{16.42} & \textbf{14.86} & 14.06 & \textbf{21.33}\\
\midrule
\multicolumn{8}{c}{\texttt{Llama-3.1-8B-instruct} on L-Eval} \\

\midrule
\textsc{FullAttn} & 10.0 & 66.84 & 38.91 & 23.11& 18.76 & 16.86 & 29.08 \\ 

\midrule
\textsc{InfLLM} & 6.67 & 54.77 & 33.76 & 20.35 & 17.62 & 16.73 & 24.98 \\ 
\textsc{HF-2bits} & 1.11 & 29.79 & 18.98 & 9.46 & 14.02 & 13.73 & 14.52\\ 
{\textsc{SirLLM}} & 5.56 & \underline{58.00} & 35.41 & \underline{21.21} & 17.32 & 16.44 & \underline{25.66} \\
\textsc{MInference} & \underline{7.78} & 31.80 & \underline{36.93} & 19.44 & \underline{18.14} & \underline{16.76} & 21.81\\
\rowcolor{cyan!10}
\textbf{\name~} & \textbf{8.89} & \textbf{63.03} & \textbf{37.21} & \textbf{23.59} & \textbf{18.17} & \textbf{16.87} & \textbf{27.96}\\
\bottomrule
\end{tabular}
}
\end{subtable}

\vspace{-10pt}
\end{table*}

\begin{figure*}[t]
\begin{center}
\includegraphics[width=0.48\linewidth]{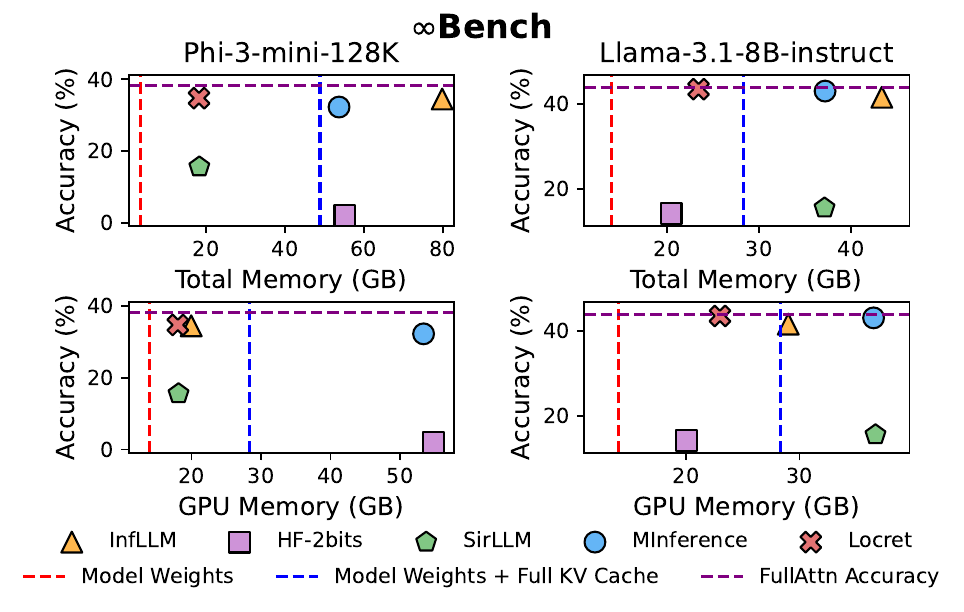}
\includegraphics[width=0.48\linewidth]{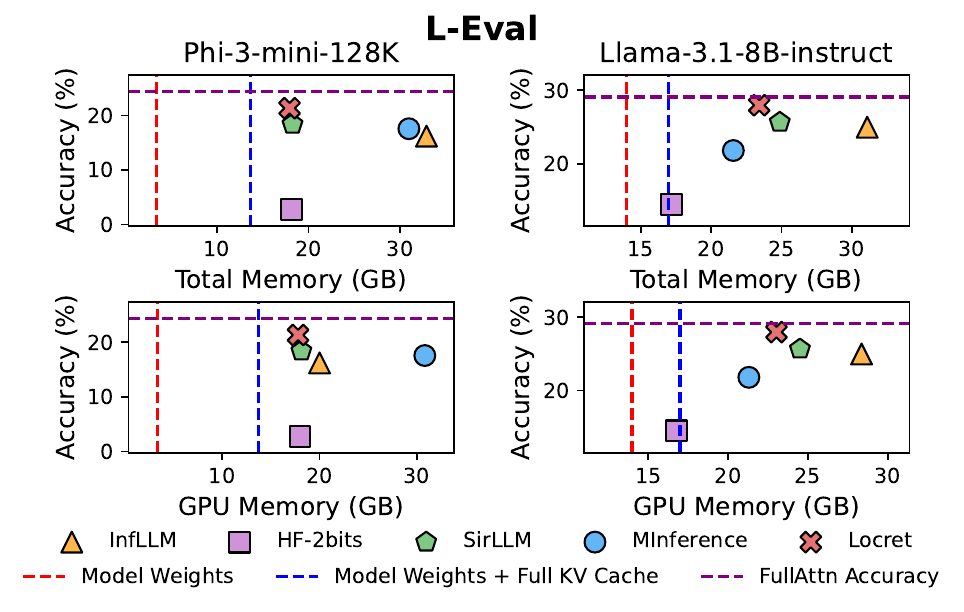}
\end{center}
\vspace{-15pt}
\caption{Memory Statistics vs. Task Performance. The red lines correspond to the theoretical size of the model weights, while the blue lines represent the total size of the model weights and the full KV cache without any compression. The purple lines indicate the accuracies of \textsc{FullAttn}. ``Total Memory'' represents the total memory usage of both GPU and CPU.}
\label{fig:memory_statics}
\vspace{-15pt}
\end{figure*}

\subsection{Experimental Setup}
\label{sec:setup}

\textbf{Model and Dataset Settings.} 
We conduct experiments on two long-context LLMs: \texttt{Phi-3-mini-128K}~\citep{abdin2024phi} and \texttt{Llama-3.1-8B-instruct}~\citep{dubey2024llama3}. 
Both models can process up to 128K context tokens and follow MHA and GQA architectures, respectively. 
The parameter sizes of these two models are also suitable for deployment on consumer-grade devices.
We inject retaining heads $\mathbf{R}$ into each layer of these two models, and the intermediate size $d_{\mathbf{R}}$ is 1,024. 
The retaining heads are trained on the LongAlpaca dataset~\citep{chen2023longlora} for 3,000 steps
\rev{, with a 5e-4 learning rate, 10,240 sequence length, and $\alpha$ set to 2.5e-3}.
Training \name~is lightweight, 
with the tunable parameters comprising 8\% and 2.5\% of the total for the two models, respectively. 
The complete training process takes 0.47 and 0.80 GPU hours on an A800 GPU for each corresponding model.
More important hyperparameters are listed in Table~\ref{tab:inf-hyper}.
More details on hyperparameters and system environments can be found in Appendix \ref{sec:hyper}.

\textbf{Benchmarks.} 
We evaluate \name~on selected subsets of $\infty$Bench~\citep{zhang2024infty} and L-Eval~\citep{an2023eval}. 
For $\infty$Bench, we select R.PassKey, R.Number, E.Sum, E.QA, E.MC, Z.QA, E.Dia, C.Debug, and M.Find. 
All selected subsets, except Z.QA, have an average length of approximately 100K tokens, while Z.QA has a longer average length of around 2000K tokens. 
We exclude R.KV because it can be easily handled by calling a Python interpreter. 
We also exclude C.Run and M.Calc due to their complexity for all methods, including full attention inference. 
For L-Eval, we filter out all tasks with an average length shorter than 16K tokens and evaluate models on CodeU, NQ, CUAD, NarrativeQA, QMSum, and SPACE. 
Metrics are reported according to the recommendations of the two datasets, with further details provided in Appendix \ref{sec:hyper}. 
We also report the peak memory usage, i.e. the average peak memory measured for the first entry of each task in $\infty$Bench and L-Eval, for reference.
\rev{
Moreover, we evaluate \name~on an extremely long-context dataset, R.PassKey with 10M tokens, in Appendix~\ref{append:extreme}. The experiments under the multi-turn conversation setting are in Appendix~\ref{append:multi-turn}.
}

\textbf{Baselines.} 
As discussed in Section~\ref{sec:related_work}, existing algorithms for memory-efficient long-context inference can be categorized into offloading-based, sparsity-based, quantization-based, and token-dropping methods. 
For each category, we select one representative method as the baseline. 
We compare \name~against full attention inference (denoted as {\textsc{FullAttn}}), \textsc{InfLLM}~\citep{xiao2024infllm}, \textsc{MInference}~\citep{jiang2024minference}, KV cache quantization~\citep{hf-quant}, and \textsc{SirLLM}~\citep{yao2024sirllm}. 
For quantization, we use HuggingFace Quanto~\citep{quanto} implementation, referring to the 2-bit quantization method as \textsc{HF-2bits}. 
We omit \textsc{HF-4bits} and benchmark the combination with \name~in Section~\ref{sec:quant}. 
We do not include attention pooling-based token-dropping methods in this benchmark, as they are orthogonal to our approach, and further discussion about this is provided in Section~\ref{sec:pool}. 
Detailed introductions to the selected baselines can be found in Appendix~\ref{sec:hyper}. 
We also discuss the comparison between the trained \name~and the randomly initialized retaining heads $\mathbf{R}$ in Appendix~\ref{sec:random}.

\subsection{End-to-end Benchmark}
\label{sec:benchmark}

We compare all the methods on $\infty$Bench and L-Eval to address \underline{\textbf{Q1}}. 
In Table~\ref{tab:results}, \name~outperforms  baselines in terms of end-to-end performance, showing:

(1) On $\infty$Bench, while all methods experience performance degradation compared to \textsc{FullAttn}, \name, \textsc{InfLLM}, and \textsc{MInference} exhibit better performance than other methods, with only a modest drop in performance given the reduced memory usage. 
Quantization shows significant degradation and fails on all tasks. 
\textsc{SirLLM} performs well on comprehensive tasks such as E.Sum and E.MC, but struggles with tasks that require precise memory, such as R.PassKey and R.Number. 
\name~not only excels in context retrieval tasks but also achieves strong results in comprehensive tasks, earning the highest overall score among all competitors.

(2) On L-Eval, all methods show performance degradation. 
Nevertheless, \name~achieves the best overall performance, obtaining the highest scores on most tasks. 
L-Eval is a shorter but more complex dataset, where \textsc{SirLLM} performs particularly well. 
Quantization fails on most tasks. 
Both \textsc{InfLLM} and \textsc{MInference} suffer significant performance drops compared to \textsc{FullAttn} inference. 
\name~consistently surpasses all competitors.

We report memory consumption in Figure~\ref{fig:memory_statics}, showing:

(1) In the extreme long-context scenario ($\infty$Bench), \name~uses relatively less memory while achieving the best overall performance. 
\textsc{InfLLM} performs well with limited GPU memory usage, but it requires a significant amount of CPU memory to store the full KV cache. 
\textsc{HF-2Bits} and \textsc{SirLLM} can achieve low memory consumption in some settings, but quantization introduces severe performance degradation. 
\textsc{MInference} employs sparse attention patterns but does not compress the KV cache. 
As a result, its minimum memory requirement equals the sum of the model weights and the full KV cache. 
In the shorter context scenario (L-Eval), a similar phenomenon is observed. 

(2) For \texttt{Phi-3-mini-128K}, which has a larger KV cache, \textsc{InfLLM} and \textsc{MInference} exhibit higher memory consumption since they store the full KV cache. 
Although other methods have similar memory footprints, \name~uses the least memory and achieves the best overall performance. 

(3) For \texttt{Llama-3.1-8B-instruct}, whose full KV cache is smaller, the memory bottleneck shifts to the runtime computational memory for attention and other calculations. 
All methods exhibit similar memory footprints, with \name~delivering the best overall performance. 

In summary, our experiments demonstrate that \name~is both effective and efficient, outperforming all baselines on multiple datasets and models while using less GPU memory.

\begin{table*}[!ht]
\small
\caption{Executing R.PassKey on an NVIDIA 4090. ``tok/s'' represents the inference speed, ``C.Len'' stands for the context length after truncation, and ``Acc.'' represents task accuracy. 
The highest score among 128K context is marked in \textbf{bold}.}
\label{tab:speed-test}
\begin{center}
\scalebox{0.84}{
\begin{tabular}{cl|ccccc>{\columncolor{cyan!10}}c|cc}
\toprule
\multicolumn{2}{c|}{Method} & \textsc{FullAttn} & \textsc{InfLLM} & \textsc{HF-2bits} & \textsc{SirLLM} &  \textsc{MInference} &  \textbf{\name} & \textsc{HF-2bits}$^*$ & \textsc{MInference}$^*$ \\ \midrule

\multirow{3}{*}{\makecell{Phi-3-\\mini-128K}} & tok/s$\uparrow$ & - & 2276.38 & - & 2352.20 & - & \textbf{5080.85} & 1098.51 & 4099.92\\

& C.Len.$\uparrow$ & 128K & 128K & 128K & 128K & 128K & \textbf{128K} & 30K  & 14K\\
& Acc.$\uparrow$ & OOM & 99.83 & OOM & 1.69 & OOM & \textbf{100.00} & 0.00 & 13.56\\
\midrule

\multirow{3}{*}{\makecell{Llama-3.1-\\8B-instruct}} & tok/s$\uparrow$ & -&  2287.66 & 1365.51 & 1589.75& -  & \textbf{3209.10}  & 3680.06 & {5135.74} \\
& C.Len.$\uparrow$ & 128K & 128K & 128K & 128K & 128K & \textbf{128K} & 30K & 25K\\
& Acc.$\uparrow$ & OOM & 100.00 & 35.59 & 1.69 & OOM & \textbf{100.00} & 26.78 & 20.34\\
\bottomrule
\end{tabular}
}
\end{center}
\vspace{-15pt}
\end{table*}

\subsection{Processing Speed on Real Consumer-Grade Devices}
\label{sec:speed_test}

We examine the processing speed to demonstrate that \name~achieves its strong performance without compromising inference speed, addressing question \underline{\textbf{Q2}}. We evaluate the inference speed on the R.PassKey task from $\infty$Bench and compare \name~against all the baselines, using a single NVIDIA 4090 GPU with 24GB of memory, which is typical for consumer-grade AI devices. 
We report the inference speed (the total number of tokens within the input and output sequences divided by the processing time) and the task accuracy. 
Since the original settings of some baselines might lead to out-of-memory (OOM) errors, we remove some tokens from the middle of the input sequence in those cases, marking these settings with $^*$, and report the valid context length in such scenario. 
For settings without $^*$, we maximize the chunk size for higher speed when the method utilizes the chunked prefill pattern.

R.PassKey is a task where the model retrieves a 5-digit number from a large amount of irrelevant text, a task we believe to be relatively simple for humans. 
Thus, we consider the task to have failed if the accuracy falls below 95\%. 
As shown in Table~\ref{tab:speed-test}, aside from the settings that fail on this task, \name~achieves the highest inference speed among all methods that can correctly process R.PassKey:

(1) Due to its MHA architecture, \texttt{Phi-3-mini-128K} has a larger KV cache, which leads to failures for both \textsc{HF-2bits} and \textsc{MInference}. 
Storing the full KV cache on a single 4090 GPU is infeasible, as it requires 48GB of memory. 
Although the quantized KV cache is reduced to 6GB, converting representations requires significant GPU memory for its intermediate states, resulting in the failure of \textsc{HF-2bits}. 
While \textsc{InfLLM} can run in memory-limited scenarios, its offloading process slows down inference. 
\textsc{SirLLM} fails due to its inaccurate eviction, which cannot correctly identify the 5-digit number.

(2) In the GQA model (\texttt{Llama-3.1-8B-instruct}), which has a smaller KV cache, the quantized cache can fit within the GPU memory. 
However, the quantization and dequantization processes become the bottleneck, leading to significantly slower speed performance. 
The performance of \textsc{InfLLM}, \textsc{SirLLM}, and \textsc{MInference} is similar to that seen with \texttt{Phi-3-mini-128K}. 
Although \textsc{MInference} benefits from faster encoding speed, it fails this task because it cannot process the entire input sequence at once. 
\name~strikes a balance between inference speed and performance, making it a far more suitable solution for long-context scenarios on consumer-grade devices.

\subsection{\nameq: Supporting Query-Driven Tasks}

We mainly address question \underline{\textbf{Q3}} in this section.
Query-driven tasks are characterized by highly sparse yet query-correlated critical regions within the context. A representative example is the Multikey-NIAH task from RULER~\citep{hsieh2024ruler}, where the context comprises JSON key-value pairs given by UUIDs, and the objective is to retrieve specific values using given keys. Such contexts are inherently challenging to compress effectively without query information.
As introduced by~\citet{sun2024shadowkv}, existing eviction-based techniques exhibit significant performance degradation when applied to query-driven tasks. Due to this, we evaluate \name~against selected eviction-based baselines using the RULER benchmark~\citep{hsieh2024ruler}.
We introduce \nameq, a query-aware variant of \name. When training the  retaining heads, we prepend the last $l_q$ query tokens to the sequence and gather CIS labels. At inference, the query is inserted at the sequence start, ensuring its visibility across all eviction actions. This adaptation enables \nameq~to perform query-aware eviction.

Table~\ref{tab:ruler} shows \nameq~against \textsc{SnapKV}~\citep{li2024snapkv}, \textsc{H$_2$O}~\citep{zhang2024h2o}, and \textsc{SirLLM}~\citep{yao2024sirllm} on RULER with a 128K context length. 
For reference, we also include results for \textsc{FullAttn} and \textsc{MInference}. Metrics include prefill speed, decode speed, and task performances are reported.
As shown in Table~\ref{tab:ruler}, all efficient inference methods exhibit performance degradation to some extent. However, \nameq~outperforms other eviction-based methods and even surpasses \textsc{MInference}, demonstrating its effectiveness on query-driven tasks. \textsc{SnapKV} shows performance degradation, while \textsc{H$_2$O} and \textsc{SirLLM} fail completely on RULER. 
For speedup, methods combining eviction and chunked prefill (\nameq, \name, and \textsc{SirLLM}) significantly reduce prefill time, achieving $>2\times$ speedup over \textsc{FullAttn}. \textsc{SnapKV} cannot accelerate prefill due to no computation reduction. \textsc{H$_2$O} suffers from extremely slow prefill as it relies on full-sequence attention scores, incompatible with efficient implementations like \textsc{Flash-Attention}. 
Decoding speeds are similar across eviction-based methods, as they maintain comparable KV cache sizes after eviction, and all of them are faster than \textsc{FullAttn}. Notably, \name~fails on RULER, showing a gap compared to \nameq, highlighting the necessity of query-awareness for query-centric tasks. A simple modification to \name~unlocks its potential for such tasks.

\begin{table}[t]
\small
\caption{Performance, prefill speed, and decode speed on RULER. The best and second-highest scores among eviction-based methods in each column are highlighted in \textbf{bold} and \underline{underlined}. ``$^\dag$'' indicates only testing the first 20 entries per subtask due to poor performance. \textsc{FullAttn} is implemented using \textsc{Flash-Attention}.}
\label{tab:ruler}
\begin{center}
\scalebox{0.84}{
\begin{tabular}{l|ccc}\toprule
    Method & RULER-128K (\%) & Prefill (tok/s) & Decode (tok/s)  \\\midrule
    \textsc{FullAttn} & 82.20 & 4319.95 & 12.32\\
    \textsc{MInference} & 72.97 & 7205.06 & 2.61\\
    \bottomrule
    \rowcolor{cyan!10}
    \nameq & \textbf{75.54} & \underline{9587.84} &{40.11}\\
    \textsc{SnapKV} & \underline{48.76} & 4203.34 & 36.49 \\
    \textsc{H$_2$O}$^\dag$ & 15.04 & 464.73 & \textbf{44.70}\\
    \textsc{SirLLM}$^\dag$ & 13.23 & \textbf{9717.41} & \underline{40.86}\\
    \name$^\dag$ & 34.33 & 9587.09 & 37.38 \\
    \bottomrule
\end{tabular}
}
\end{center}
\vspace{-25pt}
\end{table}

\vspace{-5pt}

\subsection{Additional Experiments}
Additional experiments that further investigate \name~are included in the appendices due to space limitations. We compare \name~with \textit{randomized eviction} in Appendix~\ref{sec:random}, evaluate \name~on \textit{LongBench} in Appendix~\ref{sec:longbench}, provide a \textit{hyperparameter analysis} in Appendix~\ref{sec:hyperanal}, and explore the \textit{combination of \name~with other methods} in Appendix~\ref{sec:ortho}. Please refer to the appendices for further details.



\vspace{-5pt}
\section{Conclusion}
\vspace{-1pt}

We propose \name, a lightweight training-based method that enables memory-efficient long-context LLM inference on consumer-grade devices. 
\name~introduces retaining heads to predict the CIS of each cache unit during chunked prefill and performs accurate cache eviction. 
We conduct extensive experiments across different models and multiple datasets to compare \name~with major efficient inference techniques, 
and results show that \name~outperforms all baselines, using less GPU memory and without requiring offloading to CPU memory. 
\nameq, a query-aware variant of \name, can further process query-centric tasks without significant performance degradation.
Future work will involve testing \name~on other model architectures, such as encoder-decoder and multi-latent models. More evaluations of \name~on other popular devices like NVIDIA Jetson are also planned.
We also plan to explore integrating existing KV cache budget allocation methods with \name~to further enhance inference efficiency.

\section*{Impact Statement}

This paper presents work whose goal is to advance the field of 
Machine Learning. There are many potential societal consequences 
of our work, none which we feel must be specifically highlighted here.

\bibliography{example_paper}

\begin{thebibliography}{73}
\providecommand{\natexlab}[1]{#1}
\providecommand{\url}[1]{\texttt{#1}}
\expandafter\ifx\csname urlstyle\endcsname\relax
  \providecommand{\doi}[1]{doi: #1}\else
  \providecommand{\doi}{doi: \begingroup \urlstyle{rm}\Url}\fi

\bibitem[Abdin et~al.(2024)Abdin, Jacobs, Awan, Aneja, Awadallah, Awadalla, Bach, Bahree, Bakhtiari, Behl, et~al.]{abdin2024phi}
Abdin, M., Jacobs, S.~A., Awan, A.~A., Aneja, J., Awadallah, A., Awadalla, H., Bach, N., Bahree, A., Bakhtiari, A., Behl, H., et~al.
\newblock Phi-3 technical report: A highly capable language model locally on your phone.
\newblock \emph{arXiv:2404.14219}, 2024.

\bibitem[Agrawal et~al.(2023)Agrawal, Panwar, Mohan, Kwatra, Gulavani, and Ramjee]{agrawal2023sarathi}
Agrawal, A., Panwar, A., Mohan, J., Kwatra, N., Gulavani, B.~S., and Ramjee, R.
\newblock Sarathi: Efficient llm inference by piggybacking decodes with chunked prefills.
\newblock \emph{arXiv:2308.16369}, 2023.

\bibitem[Ainslie et~al.(2023)Ainslie, Lee-Thorp, de~Jong, Zemlyanskiy, Lebr{\'o}n, and Sanghai]{ainslie2023gqa}
Ainslie, J., Lee-Thorp, J., de~Jong, M., Zemlyanskiy, Y., Lebr{\'o}n, F., and Sanghai, S.
\newblock Gqa: Training generalized multi-query transformer models from multi-head checkpoints.
\newblock \emph{Proceedings of EMNLP}, 2023.

\bibitem[An et~al.(2024)An, Gong, Zhong, Zhao, Li, Zhang, Kong, and Qiu]{an2023eval}
An, C., Gong, S., Zhong, M., Zhao, X., Li, M., Zhang, J., Kong, L., and Qiu, X.
\newblock L-eval: Instituting standardized evaluation for long context language models.
\newblock \emph{Proceedings of ACL}, 2024.

\bibitem[Anthropic(2024)]{claude3}
Anthropic.
\newblock The claude 3 model family: Opus, sonnet, haiku, 2024.
\newblock URL \url{https://www-cdn.anthropic.com/de8ba9b01c9ab7cbabf5c33b80b7bbc618857627/Model_Card_Claude_3.pdf}.

\bibitem[Bai et~al.(2024{\natexlab{a}})Bai, Lv, Zhang, He, Qi, Hou, Tang, Dong, and Li]{bai2024longalign}
Bai, Y., Lv, X., Zhang, J., He, Y., Qi, J., Hou, L., Tang, J., Dong, Y., and Li, J.
\newblock Longalign: A recipe for long context alignment of large language models.
\newblock \emph{Proceedings of EMNLP}, 2024{\natexlab{a}}.

\bibitem[Bai et~al.(2024{\natexlab{b}})Bai, Lv, Zhang, Lyu, Tang, Huang, Du, Liu, Zeng, Hou, et~al.]{bai2023longbench}
Bai, Y., Lv, X., Zhang, J., Lyu, H., Tang, J., Huang, Z., Du, Z., Liu, X., Zeng, A., Hou, L., et~al.
\newblock Longbench: A bilingual, multitask benchmark for long context understanding.
\newblock \emph{Proceedings of ACL}, 2024{\natexlab{b}}.

\bibitem[Cai et~al.(2024{\natexlab{a}})Cai, Tian, Wang, and Chen]{cai2024lococo}
Cai, R., Tian, Y., Wang, Z., and Chen, B.
\newblock Lococo: Dropping in convolutions for long context compression.
\newblock \emph{Proceedings of ICML}, 2024{\natexlab{a}}.

\bibitem[Cai et~al.(2024{\natexlab{b}})Cai, Zhang, Gao, Liu, Liu, Lu, Xiong, Dong, Chang, Hu, et~al.]{cai2024pyramidkv}
Cai, Z., Zhang, Y., Gao, B., Liu, Y., Liu, T., Lu, K., Xiong, W., Dong, Y., Chang, B., Hu, J., et~al.
\newblock Pyramidkv: Dynamic kv cache compression based on pyramidal information funneling.
\newblock \emph{arXiv:2406.02069}, 2024{\natexlab{b}}.

\bibitem[Chen et~al.(2024)Chen, Qian, Tang, Lai, Liu, Han, and Jia]{chen2023longlora}
Chen, Y., Qian, S., Tang, H., Lai, X., Liu, Z., Han, S., and Jia, J.
\newblock Longlora: Efficient fine-tuning of long-context large language models.
\newblock \emph{Proceedings of ICLR}, 2024.

\bibitem[Dao(2024)]{dao2023flashattention2}
Dao, T.
\newblock Flashattention-2: Faster attention with better parallelism and work partitioning.
\newblock \emph{Proceedings of ICLR}, 2024.

\bibitem[Dao et~al.(2022)Dao, Fu, Ermon, Rudra, and R{\'e}]{dao2022flashattention}
Dao, T., Fu, D., Ermon, S., Rudra, A., and R{\'e}, C.
\newblock Flashattention: Fast and memory-efficient exact attention with io-awareness.
\newblock \emph{Proceedings of NeurIPS}, 2022.

\bibitem[Dettmers et~al.(2022)Dettmers, Lewis, Belkada, and Zettlemoyer]{dettmers2022gpt3}
Dettmers, T., Lewis, M., Belkada, Y., and Zettlemoyer, L.
\newblock Llm.int8(): 8-bit matrix multiplication for transformers at scale.
\newblock \emph{Proceedings of NeurIPS}, 2022.

\bibitem[Dubey et~al.(2024)Dubey, Jauhri, Pandey, Kadian, Al-Dahle, Letman, Mathur, Schelten, Yang, Fan, et~al.]{dubey2024llama3}
Dubey, A., Jauhri, A., Pandey, A., Kadian, A., Al-Dahle, A., Letman, A., Mathur, A., Schelten, A., Yang, A., Fan, A., et~al.
\newblock The llama 3 herd of models.
\newblock \emph{arXiv:2407.21783}, 2024.

\bibitem[Frantar et~al.(2023)Frantar, Ashkboos, Hoefler, and Alistarh]{frantar2022gptq}
Frantar, E., Ashkboos, S., Hoefler, T., and Alistarh, D.
\newblock Gptq: Accurate post-training quantization for generative pre-trained transformers.
\newblock \emph{Proceedings of ICLR}, 2023.

\bibitem[Ge et~al.(2024)Ge, Zhang, Liu, Zhang, Han, and Gao]{ge2023model}
Ge, S., Zhang, Y., Liu, L., Zhang, M., Han, J., and Gao, J.
\newblock Model tells you what to discard: Adaptive kv cache compression for llms.
\newblock \emph{Proceedings of ICLR}, 2024.

\bibitem[Ghorpade et~al.(2012)Ghorpade, Parande, Kulkarni, and Bawaskar]{ghorpade2012gpgpu}
Ghorpade, J., Parande, J., Kulkarni, M., and Bawaskar, A.
\newblock Gpgpu processing in cuda architecture.
\newblock \emph{Proceedings of ACIJ}, 2012.

\bibitem[Hooper et~al.(2024)Hooper, Kim, Mohammadzadeh, Mahoney, Shao, Keutzer, and Gholami]{hooper2024kvquant}
Hooper, C., Kim, S., Mohammadzadeh, H., Mahoney, M.~W., Shao, Y.~S., Keutzer, K., and Gholami, A.
\newblock Kvquant: Towards 10 million context length llm inference with kv cache quantization.
\newblock \emph{arXiv:2401.18079}, 2024.

\bibitem[Hsieh et~al.(2024)Hsieh, Sun, Kriman, Acharya, Rekesh, Jia, Zhang, and Ginsburg]{hsieh2024ruler}
Hsieh, C.-P., Sun, S., Kriman, S., Acharya, S., Rekesh, D., Jia, F., Zhang, Y., and Ginsburg, B.
\newblock Ruler: What's the real context size of your long-context language models?
\newblock \emph{arXiv:2404.06654}, 2024.

\bibitem[Hu et~al.(2024{\natexlab{a}})Hu, Huang, Xu, Chen, Xu, Chen, Feng, Wang, Wang, Bao, et~al.]{hu2024inference}
Hu, C., Huang, H., Xu, L., Chen, X., Xu, J., Chen, S., Feng, H., Wang, C., Wang, S., Bao, Y., et~al.
\newblock Inference without interference: Disaggregate llm inference for mixed downstream workloads.
\newblock \emph{arXiv:2401.11181}, 2024{\natexlab{a}}.

\bibitem[Hu et~al.(2024{\natexlab{b}})Hu, Tu, Han, He, Cui, Long, Zheng, Fang, Huang, Zhao, et~al.]{hu2024minicpm}
Hu, S., Tu, Y., Han, X., He, C., Cui, G., Long, X., Zheng, Z., Fang, Y., Huang, Y., Zhao, W., et~al.
\newblock Minicpm: Unveiling the potential of small language models with scalable training strategies.
\newblock \emph{Proceedings of COLM}, 2024{\natexlab{b}}.

\bibitem[Hugging-Face()]{quanto}
Hugging-Face.
\newblock URL \url{https://github.com/huggingface/optimum-quanto}.

\bibitem[Jiang et~al.(2024{\natexlab{a}})Jiang, Li, Zhang, Wu, Luo, Ahn, Han, Abdi, Li, Lin, et~al.]{jiang2024minference}
Jiang, H., Li, Y., Zhang, C., Wu, Q., Luo, X., Ahn, S., Han, Z., Abdi, A.~H., Li, D., Lin, C.-Y., et~al.
\newblock Minference 1.0: Accelerating pre-filling for long-context llms via dynamic sparse attention.
\newblock \emph{Proceedings of ICML}, 2024{\natexlab{a}}.

\bibitem[Jiang et~al.(2024{\natexlab{b}})Jiang, Yan, Yao, Zhou, Chen, and Yuan]{jianghexgen}
Jiang, Y., Yan, R., Yao, X., Zhou, Y., Chen, B., and Yuan, B.
\newblock Hexgen: Generative inference of large language model over heterogeneous environment.
\newblock \emph{Proceedings of ICML}, 2024{\natexlab{b}}.

\bibitem[Kang et~al.(2024)Kang, Zhang, Kundu, Jeong, Liu, Krishna, and Zhao]{kang2024gear}
Kang, H., Zhang, Q., Kundu, S., Jeong, G., Liu, Z., Krishna, T., and Zhao, T.
\newblock Gear: An efficient kv cache compression recipefor near-lossless generative inference of llm.
\newblock \emph{arXiv:2403.05527}, 2024.

\bibitem[KVCache.AI(2024)]{ktransformers}
KVCache.AI.
\newblock Ktransformers: A flexible framework for experiencing cutting-edge llm inference optimizations, 2024.
\newblock URL \url{https://github.com/kvcache-ai/ktransformers}.

\bibitem[Kwon et~al.(2023)Kwon, Li, Zhuang, Sheng, Zheng, Yu, Gonzalez, Zhang, and Stoica]{kwon2023efficient}
Kwon, W., Li, Z., Zhuang, S., Sheng, Y., Zheng, L., Yu, C.~H., Gonzalez, J., Zhang, H., and Stoica, I.
\newblock Efficient memory management for large language model serving with pagedattention.
\newblock \emph{Proceedings of SOSP}, 2023.

\bibitem[Lee et~al.(2024)Lee, Lee, Seo, and Sim]{lee2024infinigen}
Lee, W., Lee, J., Seo, J., and Sim, J.
\newblock Infinigen: Efficient generative inference of large language models with dynamic kv cache management.
\newblock \emph{Proceedings of OSDI}, 2024.

\bibitem[Li et~al.(2024{\natexlab{a}})Li, Huang, Yang, Venkitesh, Locatelli, Ye, Cai, Lewis, and Chen]{li2024snapkv}
Li, Y., Huang, Y., Yang, B., Venkitesh, B., Locatelli, A., Ye, H., Cai, T., Lewis, P., and Chen, D.
\newblock Snapkv: Llm knows what you are looking for before generation.
\newblock \emph{arXiv:2404.14469}, 2024{\natexlab{a}}.

\bibitem[Li et~al.(2024{\natexlab{b}})Li, Liang, Lyu, and Wang]{li2024making}
Li, Y., Liang, S., Lyu, M.~R., and Wang, L.
\newblock Making long-context language models better multi-hop reasoners.
\newblock \emph{Proceedings of ACL}, 2024{\natexlab{b}}.

\bibitem[Lin et~al.(2024)Lin, Tang, Tang, Yang, Chen, Wang, Xiao, Dang, Gan, and Han]{lin2024awq}
Lin, J., Tang, J., Tang, H., Yang, S., Chen, W.-M., Wang, W.-C., Xiao, G., Dang, X., Gan, C., and Han, S.
\newblock Awq: Activation-aware weight quantization for on-device llm compression and acceleration.
\newblock \emph{Proceedings of MLSys}, 2024.

\bibitem[Liu et~al.(2024{\natexlab{a}})Liu, Desai, Liao, Wang, Xie, Xu, Kyrillidis, and Shrivastava]{liu2024scissorhands}
Liu, Z., Desai, A., Liao, F., Wang, W., Xie, V., Xu, Z., Kyrillidis, A., and Shrivastava, A.
\newblock Scissorhands: Exploiting the persistence of importance hypothesis for llm kv cache compression at test time.
\newblock \emph{Proceedings of NeurIPS}, 2024{\natexlab{a}}.

\bibitem[Liu et~al.(2024{\natexlab{b}})Liu, Yuan, Jin, Zhong, Xu, Braverman, Chen, and Hu]{liu2024kivi}
Liu, Z., Yuan, J., Jin, H., Zhong, S., Xu, Z., Braverman, V., Chen, B., and Hu, X.
\newblock Kivi: A tuning-free asymmetric 2bit quantization for kv cache.
\newblock \emph{Proceedings of ICML}, 2024{\natexlab{b}}.

\bibitem[llama2.c()]{llama2c}
llama2.c.
\newblock URL \url{https://github.com/karpathy/llama2.c}.

\bibitem[llama.cpp()]{llamacpp}
llama.cpp.
\newblock URL \url{https://github.com/ggerganov/llama.cpp}.

\bibitem[Loshchilov(2019)]{loshchilov2017decoupled}
Loshchilov, I.
\newblock Decoupled weight decay regularization.
\newblock \emph{Proceedings of ICLR}, 2019.

\bibitem[Lou et~al.(2024)Lou, Jia, Zheng, and Tu]{lou2024sparser}
Lou, C., Jia, Z., Zheng, Z., and Tu, K.
\newblock Sparser is faster and less is more: Efficient sparse attention for long-range transformers.
\newblock \emph{arXiv:2406.16747}, 2024.

\bibitem[Lu et~al.(2024)Lu, Li, Cai, Yi, Liu, Zhang, Lane, and Xu]{lu2024small}
Lu, Z., Li, X., Cai, D., Yi, R., Liu, F., Zhang, X., Lane, N.~D., and Xu, M.
\newblock Small language models: Survey, measurements, and insights.
\newblock \emph{arXiv:2409.15790}, 2024.

\bibitem[Lv et~al.(2024)Lv, Feng, Xie, Jia, Peng, and Xie]{lv2024critiprefill}
Lv, J., Feng, Y., Xie, X., Jia, X., Peng, Q., and Xie, G.
\newblock Critiprefill: A segment-wise criticality-based approach for prefilling acceleration in llms.
\newblock \emph{arXiv:2409.12490}, 2024.

\bibitem[Mei et~al.(2024)Mei, Li, Xu, Ye, Ge, and Zhang]{mei2024aios}
Mei, K., Li, Z., Xu, S., Ye, R., Ge, Y., and Zhang, Y.
\newblock Aios: Llm agent operating system.
\newblock \emph{arXiv:2403.16971}, 2024.

\bibitem[Minaee et~al.(2024)Minaee, Mikolov, Nikzad, Chenaghlu, Socher, Amatriain, and Gao]{minaee2024large}
Minaee, S., Mikolov, T., Nikzad, N., Chenaghlu, M., Socher, R., Amatriain, X., and Gao, J.
\newblock Large language models: A survey.
\newblock \emph{arXiv:2402.06196}, 2024.

\bibitem[Mu et~al.(2024)Mu, Li, and Goodman]{mu2024learning}
Mu, J., Li, X., and Goodman, N.
\newblock Learning to compress prompts with gist tokens.
\newblock \emph{Proceedings of NeurIPS}, 2024.

\bibitem[Munkhdalai et~al.(2024)Munkhdalai, Faruqui, and Gopal]{munkhdalai2024leave}
Munkhdalai, T., Faruqui, M., and Gopal, S.
\newblock Leave no context behind: Efficient infinite context transformers with infini-attention.
\newblock \emph{arXiv:2404.07143}, 2024.

\bibitem[Nawrot et~al.(2024)Nawrot, {\L}a{\'n}cucki, Chochowski, Tarjan, and Ponti]{nawrot2024dynamic}
Nawrot, P., {\L}a{\'n}cucki, A., Chochowski, M., Tarjan, D., and Ponti, E.~M.
\newblock Dynamic memory compression: Retrofitting llms for accelerated inference.
\newblock \emph{arXiv:2403.09636}, 2024.

\bibitem[OpenAI(2024)]{gpt4o}
OpenAI.
\newblock Openai gpt-4o, 2024.
\newblock URL \url{https://platform.openai.com/docs/models/gpt-4o}.

\bibitem[Pan(2024)]{anti-haystack}
Pan, W.
\newblock Anti-haystack, 2024.
\newblock URL \url{https://huggingface.co/datasets/wenbopan/anti-haystack}.

\bibitem[Qin et~al.(2024{\natexlab{a}})Qin, Li, He, Zhang, Wu, Zheng, and Xu]{qin2024mooncake}
Qin, R., Li, Z., He, W., Zhang, M., Wu, Y., Zheng, W., and Xu, X.
\newblock Mooncake: Kimi's kvcache-centric architecture for llm serving.
\newblock \emph{arXiv:2407.00079}, 2024{\natexlab{a}}.

\bibitem[Qin et~al.(2024{\natexlab{b}})Qin, Hu, Lin, Chen, Ding, Cui, Zeng, Huang, Xiao, Han, Fung, Su, Wang, Qian, Tian, Zhu, Liang, Shen, Xu, Zhang, Ye, Li, Tang, Yi, Zhu, Dai, Yan, Cong, Lu, Zhao, Huang, Yan, Han, Sun, Li, Phang, Yang, Wu, Ji, Liu, and Sun]{qin2023tool}
Qin, Y., Hu, S., Lin, Y., Chen, W., Ding, N., Cui, G., Zeng, Z., Huang, Y., Xiao, C., Han, C., Fung, Y.~R., Su, Y., Wang, H., Qian, C., Tian, R., Zhu, K., Liang, S., Shen, X., Xu, B., Zhang, Z., Ye, Y., Li, B., Tang, Z., Yi, J., Zhu, Y., Dai, Z., Yan, L., Cong, X., Lu, Y., Zhao, W., Huang, Y., Yan, J., Han, X., Sun, X., Li, D., Phang, J., Yang, C., Wu, T., Ji, H., Liu, Z., and Sun, M.
\newblock Tool learning with foundation models.
\newblock \emph{ACM Computing Surveys}, 2024{\natexlab{b}}.

\bibitem[Rajput et~al.(2024)Rajput, Sheng, Owen, and Chiley]{rajput2024infer}
Rajput, S., Sheng, Y., Owen, S., and Chiley, V.
\newblock Inference-friendly models with mixattention.
\newblock \emph{arXiv:2409.15012}, 2024.

\bibitem[Reid et~al.(2024)Reid, Savinov, Teplyashin, Lepikhin, Lillicrap, Alayrac, Soricut, Lazaridou, Firat, Schrittwieser, et~al.]{reid2024gemini}
Reid, M., Savinov, N., Teplyashin, D., Lepikhin, D., Lillicrap, T., Alayrac, J.-b., Soricut, R., Lazaridou, A., Firat, O., Schrittwieser, J., et~al.
\newblock Gemini 1.5: Unlocking multimodal understanding across millions of tokens of context.
\newblock \emph{arXiv:2403.05530}, 2024.

\bibitem[rustformers()]{rustformers}
rustformers.
\newblock URL \url{https://github.com/rustformers/llm}.

\bibitem[Schnitzler et~al.(2024)Schnitzler, Ho, Huang, Boudin, Sugawara, and Aizawa]{schnitzler2024morehopqa}
Schnitzler, J., Ho, X., Huang, J., Boudin, F., Sugawara, S., and Aizawa, A.
\newblock Morehopqa: More than multi-hop reasoning.
\newblock \emph{arXiv:2406.13397}, 2024.

\bibitem[Shah et~al.(2024)Shah, Bikshandi, Zhang, Thakkar, Ramani, and Dao]{shah2024flashattention}
Shah, J., Bikshandi, G., Zhang, Y., Thakkar, V., Ramani, P., and Dao, T.
\newblock Flashattention-3: Fast and accurate attention with asynchrony and low-precision.
\newblock \emph{arXiv:2407.08608}, 2024.

\bibitem[Sheng et~al.(2023)Sheng, Zheng, Yuan, Li, Ryabinin, Chen, Liang, R{\'e}, Stoica, and Zhang]{sheng2023flexgen}
Sheng, Y., Zheng, L., Yuan, B., Li, Z., Ryabinin, M., Chen, B., Liang, P., R{\'e}, C., Stoica, I., and Zhang, C.
\newblock Flexgen: High-throughput generative inference of large language models with a single gpu.
\newblock \emph{Proceedings of ICML}, 2023.

\bibitem[Shi et~al.(2024)Shi, Zhang, Yao, Li, and Zhao]{luohe2024keep}
Shi, L., Zhang, H., Yao, Y., Li, Z., and Zhao, H.
\newblock Keep the cost down: A review on methods to optimize llm's kv-cache consumption.
\newblock \emph{Proceedings of COLM}, 2024.

\bibitem[Sun et~al.(2024)Sun, Chang, Bao, Zheng, Zheng, Liu, Dong, Chi, and Chen]{sun2024shadowkv}
Sun, H., Chang, L.-W., Bao, W., Zheng, S., Zheng, N., Liu, X., Dong, H., Chi, Y., and Chen, B.
\newblock Shadowkv: Kv cache in shadows for high-throughput long-context llm inference.
\newblock \emph{arXiv:2410.21465}, 2024.

\bibitem[Turganbay(2024)]{hf-quant}
Turganbay, R.
\newblock Unlocking longer generation with key-value cache quantization, 2024.
\newblock URL \url{https://huggingface.co/blog/kv-cache-quantization}.

\bibitem[Wang et~al.(2024)Wang, Ma, Feng, Zhang, Yang, Zhang, Chen, Tang, Chen, Lin, et~al.]{wang2024survey}
Wang, L., Ma, C., Feng, X., Zhang, Z., Yang, H., Zhang, J., Chen, Z., Tang, J., Chen, X., Lin, Y., et~al.
\newblock A survey on large language model based autonomous agents.
\newblock \emph{Frontiers of Computer Science}, 18\penalty0 (6):\penalty0 186345, 2024.

\bibitem[Wu et~al.(2024)Wu, Ren, Yang, Parasyris, Georgakoudis, Laguna, and Li]{wu2024offload}
Wu, J., Ren, J., Yang, S., Parasyris, K., Georgakoudis, G., Laguna, I., and Li, D.
\newblock Lm-offload: Performance model-guided generative inference of large language models with parallelism control.
\newblock \emph{Blog of PASA Lab}, 2024.

\bibitem[Xiao et~al.(2024{\natexlab{a}})Xiao, Zhang, Han, Xiao, Lin, Zhang, Liu, Han, and Sun]{xiao2024infllm}
Xiao, C., Zhang, P., Han, X., Xiao, G., Lin, Y., Zhang, Z., Liu, Z., Han, S., and Sun, M.
\newblock Infllm: Unveiling the intrinsic capacity of llms for understanding extremely long sequences with training-free memory.
\newblock \emph{Proceedings of NeurIPS}, 2024{\natexlab{a}}.

\bibitem[Xiao et~al.(2023)Xiao, Lin, Seznec, Wu, Demouth, and Han]{xiao2023smoothquant}
Xiao, G., Lin, J., Seznec, M., Wu, H., Demouth, J., and Han, S.
\newblock Smoothquant: Accurate and efficient post-training quantization for large language models.
\newblock \emph{Proceedings of ICML}, 2023.

\bibitem[Xiao et~al.(2024{\natexlab{b}})Xiao, Tang, Zuo, Guo, Yang, Tang, Fu, and Han]{xiao2024duoattention}
Xiao, G., Tang, J., Zuo, J., Guo, J., Yang, S., Tang, H., Fu, Y., and Han, S.
\newblock Duoattention: Efficient long-context llm inference with retrieval and streaming heads.
\newblock \emph{arXiv:2410.10819}, 2024{\natexlab{b}}.

\bibitem[Xiao et~al.(2024{\natexlab{c}})Xiao, Tian, Chen, Han, and Lewis]{xiao2023efficient}
Xiao, G., Tian, Y., Chen, B., Han, S., and Lewis, M.
\newblock Efficient streaming language models with attention sinks.
\newblock \emph{Proceedings of ICLR}, 2024{\natexlab{c}}.

\bibitem[Yang et~al.(2024)Yang, Han, Gao, Hu, Zhang, and Zhao]{yang2024pyramidinfer}
Yang, D., Han, X., Gao, Y., Hu, Y., Zhang, S., and Zhao, H.
\newblock Pyramidinfer: Pyramid kv cache compression for high-throughput llm inference.
\newblock \emph{Proceedings of ACL}, 2024.

\bibitem[Yao et~al.(2024)Yao, Li, and Zhao]{yao2024sirllm}
Yao, Y., Li, Z., and Zhao, H.
\newblock Sirllm: Streaming infinite retentive llm.
\newblock \emph{Proceedings of ACL}, 2024.

\bibitem[Yuan et~al.(2024)Yuan, Liu, Chuang, Li, Wang, Le, Jin, Chaudhary, Xu, Liu, et~al.]{yuan2024kv}
Yuan, J., Liu, H., Chuang, Y.-N., Li, S., Wang, G., Le, D., Jin, H., Chaudhary, V., Xu, Z., Liu, Z., et~al.
\newblock Kv cache compression, but what must we give in return? a comprehensive benchmark of long context capable approaches.
\newblock \emph{Proceedings of EMNLP}, 2024.

\bibitem[Zandieh et~al.(2024)Zandieh, Daliri, and Han]{zandieh2024qjl}
Zandieh, A., Daliri, M., and Han, I.
\newblock Qjl: 1-bit quantized jl transform for kv cache quantization with zero overhead.
\newblock \emph{arXiv:2406.03482}, 2024.

\bibitem[Zhang et~al.(2024{\natexlab{a}})Zhang, Yi, Xu, and Shrivastava]{zhang2024kv}
Zhang, T., Yi, J., Xu, Z., and Shrivastava, A.
\newblock Kv cache is 1 bit per channel: Efficient large language model inference with coupled quantization.
\newblock \emph{arXiv:2405.03917}, 2024{\natexlab{a}}.

\bibitem[Zhang et~al.(2024{\natexlab{b}})Zhang, Chen, Hu, Xu, Chen, Hao, Han, Thai, Wang, Liu, et~al.]{zhang2024infty}
Zhang, X., Chen, Y., Hu, S., Xu, Z., Chen, J., Hao, M.~K., Han, X., Thai, Z.~L., Wang, S., Liu, Z., et~al.
\newblock $\infty $ bench: Extending long context evaluation beyond 100k tokens.
\newblock \emph{Proceedings of ACL}, 2024{\natexlab{b}}.

\bibitem[Zhang et~al.(2024{\natexlab{c}})Zhang, Liu, Chen, Kailkhura, Chen, and Wang]{zhang2024q}
Zhang, Z., Liu, S., Chen, R., Kailkhura, B., Chen, B., and Wang, A.
\newblock Q-hitter: A better token oracle for efficient llm inference via sparse-quantized kv cache.
\newblock \emph{Proceedings MLSys}, 2024{\natexlab{c}}.

\bibitem[Zhang et~al.(2024{\natexlab{d}})Zhang, Sheng, Zhou, Chen, Zheng, Cai, Song, Tian, R{\'e}, Barrett, et~al.]{zhang2024h2o}
Zhang, Z., Sheng, Y., Zhou, T., Chen, T., Zheng, L., Cai, R., Song, Z., Tian, Y., R{\'e}, C., Barrett, C., et~al.
\newblock H2o: Heavy-hitter oracle for efficient generative inference of large language models.
\newblock \emph{Proceedings of NeurIPS}, 2024{\natexlab{d}}.

\bibitem[Zhao et~al.(2023)Zhao, Zhou, Li, Tang, Wang, Hou, Min, Zhang, Zhang, Dong, et~al.]{zhao2023survey}
Zhao, W.~X., Zhou, K., Li, J., Tang, T., Wang, X., Hou, Y., Min, Y., Zhang, B., Zhang, J., Dong, Z., et~al.
\newblock A survey of large language models.
\newblock \emph{arXiv:2303.18223}, 2023.

\bibitem[Zhong et~al.(2024)Zhong, Liu, Chen, Hu, Zhu, Liu, Jin, and Zhang]{zhong2024distserve}
Zhong, Y., Liu, S., Chen, J., Hu, J., Zhu, Y., Liu, X., Jin, X., and Zhang, H.
\newblock Distserve: Disaggregating prefill and decoding for goodput-optimized large language model serving.
\newblock \emph{Proceedings of OSDI}, 2024.

\end{thebibliography}
\bibliographystyle{icml2025}

\newpage
\appendix
\onecolumn

\section{Hyperparameters, Environment and Baselines}
\label{sec:hyper}

\subsection{Training}

During the training stage, we first insert retaining head $\mathbf{R}$s to each layar.
A retaining head is a small FFN consist of two linear transformations, and the non-linear function is aligned with other non-linears of the conresponding model, with an intermediate size of 1024.
We train the appended retaining head $\mathbf{R}$s on the LongAlpaca for 3000 steps with batch size set to 1 and maximum sequence length set to 10240. 
We use the AdamW scheduler~\citep{loshchilov2017decoupled} and the learning rate is set to 5e-4. 
We conduct the training with a linear learning rate scheduler, whose warmup step number is set to 2000. 
The balance factor between two training loss $\alpha$ is set to 0.0025.

\subsection{Inference}

\begin{table}[h]
\small
\caption{\rev{Hyperparameters in \name's inference stage. ``$b$'' is cache budget, ``$B$'' refers to chunk size of chunked prefill, ``$n_s$'' refers to stabilizers length and ``$n_{loc}$'' is local length.}}
\label{tab:inf-hyper}
\begin{center}
\scalebox{0.8}{
\begin{tabular}{l|cccc}\toprule
    Model & $b$ & $B$ & $n_s$ & $n_{loc}$ \\\midrule
    \texttt{Phi-3-mini-128K} & 6000 & 3072 & 2500 & 100 \\
    \texttt{Llama-3.1-8B-instruct} & 16384 & 1024 & 2500 & 100 \\\bottomrule
\end{tabular}
}
\end{center}
\vspace{-10pt}
\end{table}

The inference hyperparameters of \name~is listed in Table \ref{tab:inf-hyper}. 
Here, we follow the notations in Algorithm~\ref{alg:infer}.
$b$ stands for the cache budget, $B$ is the chunk size of chunked prefill, $n_s$ is the length of stabilizers, and $n_{loc}$ represents the length of locally retained tokens at the end of the input sequence.

Hyperparameters of other baselines are as follows. 
For \textsc{InfLLM}, we use the recommended settings for Llama-3 to evaluate Llama-3.1. 
Since there is no recommendations of \texttt{Phi-3-mini-128K}, we use the settings for MiniCPM, whose architechture and size is similar to \texttt{Phi-3-mini-128K}, to conduct all the experiments. 
For Quantization, we use the official implementation (Quanto backend) of Hugging Face.
For \textsc{SirLLM}, we set the start size to 4, recent size to 1000 for both models. 
We set the token entropy size to 6000 and 16384 for \texttt{Phi-3-mini-128K} and \texttt{Llama-3.1-8B-instruct} respectively.
The chunk size of chunked prefill is also 3072 and 1024 for the corresponding model.
For \textsc{MInference}, we utilize the recommended settings for both models.

\subsection{System Environment}

For all the experiments except the 4090 experiments in Section \ref{sec:speed_test}, we use a workstation with 8$\times$NVIDIA A800/H800 GPUs and 104 Intel(R) Xeon(R) Platinum 8470 CPUs. 
\rev{
We only use 1 GPU from the cluster for training, as the GPU requirements are less than 80GB for all training procedures.
}
The device has 1.0 TB CPU memory. The operating system is Red Hat 4.8.5. We conduct all experiments except the full attention full KV cache inference on a single GPU, and 2 GPUs for full attention settings.

For Section \ref{sec:speed_test}, we conduct the experiments on a single NVIDIA 4090 GPU. The device has 512 AMD EPYC 9754 128-Core Processors and 1.0 TB CPU memory. GPUs and CPUs are connected through PCIe Gen 4, which has 16GT/s transmission speed. The operating system is Ubuntu 9.4.0.

\subsection{Baselines}

We compare \name~with full attention inference, \textsc{InfLLM}, Quantization, \textsc{SirLLM} and \textsc{MInference}.
\textsc{FullAttn} inference is performed using vllm~\citep{kwon2023efficient}, which includes automatic tensor parallelism. 
\textsc{InfLLM} is a representative of the offloading-based methods, where the full KV cache is offloaded to CPU, and the most relavant blocks are retrieved to GPU during inference. 
For quantization method, we use the Hugging Face implementation of 2-bits KV cache quantization, which is inspired by~\citet{liu2024kivi}, where quantization is conducted along channels instead of tokens. 
We denote this method as \textsc{HF-2bits}. 
\textsc{SirLLM} is an eviction-based token dropping algorithm, where tokens with low token-entropy is evicted once the cache is fullfilled. 
\rev{
We use the official implementation of SirLLM, which includes some CPU operations including importance sorting.
}
\textsc{MInference} is a typical method of reducing peak GPU memory consumption through rule-based sparse attention, but it does not reduce the size of KV cache.  
Note that \textsc{InfLLM}, \textsc{HF-2bits} and \textsc{SirLLM} does not have official implementation on \texttt{Phi-3-mini-128K}, thus we implement these three methods according to the original algorithm. 
We only use the short factor of RoPE for \textsc{InfLLM}, and no further model modification is conducted for \textsc{HF-2bits} and \textsc{SirLLM}.

\section{\rev{The Global and Local Discrepancy of Scoring Functions}}
\label{append:discrepancy}
\rev{
Cache importance scoring functions can generally be categorized into two types: causal and non-causal.
Non-causal functions, e.g. \textsc{H$_2$O} and \textsc{SnapKV}, require information from subsequent cache units to determine the importance score of a cache unit, making them dependent on prefilling the entire sequence. 
On the other hand, causal functions, e.g. \textsc{SirLLM} and \name, predict cache importance without relying on subsequent information.
Non-causal scoring functions are incompatible with chunked prefill because they cannot calculate scores without access to the full sequence. If such functions are integrated with chunked prefill, they often face a significant discrepancy between the local importance score (without considering subsequent information) and the global importance score (with full context).}

\rev{To investigate this discrepancy, we measure the consistency of the top 10\% most important cache positions identified in prefixes of various lengths compared to the full context. 
For reference, the full context is truncated to 6K tokens. 
The results shown in Figure~\ref{fig:consistency} highlights that scoring functions requiring future information, such as \textsc{H$_2$O} and \textsc{SnapKV}, suffer from significant discrepancies when subsequent cache units are not considered. \textsc{SirLLM}, while also causal, shows notable inaccuracies, leading to performance degradation as demonstrated in Table~\ref{tab:results} and Table~\ref{tab:speed-test}.}

\rev{We also evaluate the end-to-end performance using \textsc{H$_2$O} and \textsc{SnapKV} with chunked prefill on $\infty$Bench, shown in Table~\ref{tab:scoring_functions}. The results demonstrate that discrepancies between local and global importance scores in H$_2$O and \textsc{SnapKV} lead to severe performance drops, particularly in R.Number. It is this discrepancy that leads to the failure of H$_2$O and \textsc{SnapKV} in accurately retrieving information from the context. Specifically, the model is unable to identify the importance of certain cache units at the time they are first encountered.~\name, however, avoids such inconsistencies and achieves superior performance.
}

\begin{table}[H]
\small
\vspace{-5pt}
\caption{\rev{$\infty$Bench scores of \textsc{H$_2$O}, \textsc{SnapKV} and~\name.}}
\vspace{5pt}
\label{tab:scoring_functions}
\begin{subtable}{1\textwidth}
\centering
\scalebox{0.85}{
\begin{tabular}{l|cccc|c}
\toprule
\multicolumn{6}{c}{\texttt{Phi-3-mini-128K} on $\infty$Bench} \\
\midrule
Method & R.Number & E.Sum & E.MC & C.Debug & Avg.$\uparrow$\\
\midrule

\textsc{FullAttn} & 97.12 & 17.92 & 55.46 & 23.10 & 48.40 \\ 
\midrule

\textsc{H$_2$O} & 3.39 & 15.35 & 45.41 & 20.57 & 21.18 \\
\textsc{SnapKV} & 2.54 & 15.44 & 41.92 & 21.43 & 20.33 \\
\rowcolor{cyan!10}
\textbf{\name} & \textbf{97.46} & \textbf{16.82} & \textbf{46.29} & \textbf{29.71} & \textbf{47.57} \\

\bottomrule

\end{tabular}
}
\end{subtable}

\end{table}

\section{Pseudocode of \name}

We provide the pseudocode of \name in this section, and we describe the inferece process of \name~in Algorithm~\ref{alg:infer}.

\begin{algorithm}[h]
    \small
    \caption{\name~Inference}
    \label{alg:infer}
    \begin{algorithmic}
        
    \STATE {\bfseries Input:}{Model $\mathbf{M}$, Prompt tokens $x$, Local length $n_{loc}$, Stablizer length $n_s$, Budget $b$, Chunk size $B$}
    \STATE {\bfseries Output:}{Generated tokens $x_{gen}$}
    \STATE {\bfseries // }{Leave the last $n_{loc}$ out to make sure they are not evicted.}
    \STATE chunk\_positions $\leftarrow$ split\_chunk($0$, $x$.length() $-n_{loc}$, $B$)
    \STATE K\_cache, V\_cache, score\_cache $\leftarrow$ [], [], []
    
    \FOR{\rm{chunk} $\in$ \rm{chunk\_positions}}
        \STATE begin\_pos, end\_pos $\leftarrow$ chunk.begin\_pos, chunk.end\_pos
        \STATE K\_chunk, V\_chunk, score\_chunk $\leftarrow$
        \STATE{\bfseries   \quad}$\mathbf{M}(x[\text{begin\_pos:end\_pos}], \text{K\_cache, V\_cache})$
        \STATE K\_cache $\leftarrow$ Concat(K\_cache, K\_chunk)
        \STATE V\_cache $\leftarrow$ Concat(V\_cache, V\_chunk)
        \STATE score\_cache $\leftarrow$ Concat(score\_cache, score\_chunk)
        \IF{\rm{chunk} is not the last chunk}
        \STATE {\bfseries // }{Keep the last $n_s$ caches to maintain higher context continuity.}
        \STATE score\_cache[score\_cache.length()-
        \STATE{\bfseries   \quad}$n_s$:score\_cache.length()] $\leftarrow +\infty$
        \ENDIF
        \STATE indices $\leftarrow$ top-$b$(score\_cache).indices
        \STATE K\_cache, V\_cache, score\_cache = K\_cache[indices],
        \STATE{\bfseries   \quad}V\_cache[indices], score\_cache[indices]
    \ENDFOR
    \STATE K\_cache, V\_cache, score\_cache $\leftarrow$
    \STATE{\bfseries   \quad}$ \mathbf{M}$($x$[$x$.length()$-n_{loc}$:$x$.length()], K\_cache, V\_cache)
    \STATE $x_{gen}\leftarrow \mathbf{M}$.generate(K\_cache, V\_cache)
    \STATE{\bfseries return } $x_{gen}$
    \end{algorithmic}
    
\end{algorithm}

\section{Evaluating \nameq~on RULER}

\begin{table*}[!h]
\small
\label{tab:results-ruler}

\centering

\vspace{-5pt}
\caption{Performance, prefill speed, and decode speed on RULER (Detailed). The best and second-highest scores among eviction-based methods in each column are highlighted in \textbf{bold} and \underline{underlined}, respectively. ``$^\dag$'' indicates testing limited to the first 20 entries per subtask due to poor performance. \textsc{FullAttn} is implemented using \textsc{Flash-Attention}.}
\vspace{5pt}
\scalebox{0.87}{
\begin{tabular}{l|ccccccccccccc|c}
\toprule
Method & SG1 & SG2 & SG3 & MK1 & MK2 & MK3 & MV & MQ & VT & CWE & FWE & QA1 & QA2 & Avg.\\
\midrule
\textsc{FullAttn} & 
99.40 & 99.80 & 99.60 & 98.20 & 87.60 & 67.00 & 94.65 & 98.00 & 60.98 & 71.40 & 72.20 & 78.20 & 41.6 & 82.20
\\ 
\textsc{MInference} &
{100.00} & 98.60 & 99.00 & 95.40 & 58.20 & 23.80 & 84.35 & 95.70& {66.40} & 45.94 & 74.67 & 67.80 & 38.80 & 72.97
\\ \midrule
\rowcolor{cyan!10}
\nameq & \textbf{100.00} & \textbf{99.80} & \textbf{99.60} & \textbf{75.00} & \textbf{98.80} & \textbf{98.60} & \underline{66.95} & \textbf{85.50} & \underline{52.64} & 30.90 & \textbf{80.27} & \underline{53.20} & \textbf{40.80} & \textbf{75.54}\\
\textsc{SnapKV} & \textbf{100.00} & \underline{82.80} & 11.60 & \underline{70.40} & \underline{6.20} & 1.40 & \textbf{68.50} & \underline{78.40} & 49.04 & \underline{31.90} & 50.07 & 51.20 & \underline{32.40} & \underline{48.76}\\
\textsc{H$_2$O}$^\dag$ & \underline{20.00} & 0.00 & 0.00 & 0.00 & 0.00 & 0.00 & 3.75 & 2.50 & 1.00 & 0.00 & 53.33 & \textbf{100.00} & 15.00 & 15.04\\
\textsc{SirLLM}$^\dag$ & 0.00 & 5.00 & 5.00 & 0.00 & 5.00 & 5.00 & \underline{6.25} & 8.75 & 12.00 & 0.00 & \underline{80.00} & 30.00 & 15.00 & 13.23 \\
\name$^\dag$ & \textbf{100.00} & 45.00 & \underline{35.00} & 10.00 & 5.00 & 0.00 & 20.00 & 17.50 & \textbf{69.00} & \textbf{46.50} & 73.33 & 20.00 & 5.00 & 34.33 \\
\bottomrule
\end{tabular}
}
\end{table*}

To evaluate \nameq's performance on query-centric tasks, we compare it with selected eviction-based baselines: \textsc{SnapKV}, \textsc{H$_2$O}, \textsc{SirLLM}, and vanilla \name. We also include \textsc{FullAttn} (implemented with \textsc{Flash-Attention}) and \textsc{MInference} for reference. The RULER benchmark consists of 500 synthetic queries per task, each with a context length of 128K tokens. All methods are tested on \texttt{Llama-3.1-8B-instruct}. 

For \nameq~and \name, we set the budget size $b$ to 6000, chunk size $B$ to 4096, stabilizers length $n_s$ to 2500, and local length $n_{loc}$ to 100. For \textsc{SnapKV}, the voting window size is set to 100, with the last 100 tokens retained. For \textsc{H$_2$O}, due to its reliance on full-sequence attention scores, we use a layer-wise chunked prefill pattern with a chunk size of 1024. Larger chunk size would result in an out-of-memory error. For \textsc{SirLLM}, we configure the start size to 4, recent size to 1000, and budget size to 6000. All evaluations are conducted on a single NVIDIA A800-80GB GPU. 

For prefill and decode speed testing, all methods except \textsc{H$_2$O} are implemented with \textsc{Flash-Attention}; \textsc{H$_2$O} uses PyTorch's vanilla attention due to its incompatibility with efficient attention implementations.

\section{Trained Retaining Heads vs. Random Eviction}
\label{sec:random}

\begin{table}[h]
\small
\vspace{-5pt}
\caption{The results of \name~compared with randomly initialized retaining head $\mathbf{R}$s on $\infty$Bench and L-Eval.}
\vspace{5pt}
\label{tab:random}
\begin{subtable}{1\textwidth}
\centering
\scalebox{0.85}{
\begin{tabular}{l|ccccccccc|c}
\toprule
\multicolumn{11}{c}{\texttt{Phi-3-mini-128K} on $\infty$Bench} \\
\midrule
Method & R.PassKey & R.Number & E.Sum & E.QA & E.MC & Z.QA & E.Dia & C.Debug & M.Find & Avg.\\
\midrule

Random & 0.00 & 34.00 & 5.09 & 2.68 & 18.34 & 1.54 & 0.00 & 13.71 & 2.57 & 4.92\\
\rowcolor{cyan!10}
\textbf{\name~} & \textbf{100.00} & \textbf{97.46}  & \textbf{16.82}  & \textbf{7.61}  & \textbf{46.29}  & \textbf{11.31}  & \textbf{10.00}  & \textbf{27.92}  & \textbf{29.71} & \textbf{34.73}\\
\bottomrule

\end{tabular}
}
\end{subtable}
\end{table}

We compare the trained \name~to appending randomly initialized retaining head $\mathbf{R}$s on $\infty$Bench. 
The results in Table \ref{tab:random} show that \name~training is effective. 
Randomly initialized of retaining heads give random predictions and evict arbitary cache units at each step, resulting the failure on all tasks.

\section{Evaluation on LongBench}
\label{sec:longbench}

\rev{
We conduct additional experiments to evaluate Locret on LongBench~\citep{bai2023longbench}, comparing it with baselines such as Full Attention, MInference, InfLLM, and SirLLM. For this evaluation, we used \texttt{Phi-3-mini-128K} with a retained head trained on LongAlign. To ensure a fair comparison, we excluded all Chinese subtasks from LongBench and focused solely on the English subtasks, as \texttt{Phi-3-mini-128K} was not specifically trained on Chinese corpora. The results are presented below. For \name~, we follow the hyperparameters presented in Table~\ref{tab:inf-hyper}.
}

\begin{table}[H]
\small
\vspace{-5pt}
\caption{\rev{LongBench scores of \name~compared with baselines.}}
\vspace{5pt}
\label{tab:longben}
\centering
\scalebox{0.64}{
\begin{tabular}{l|cccccccccccccccc|c}
\toprule
Method & \makecell{gov\_\\report} & triviaqa & \makecell{narrative\\qa} & qmsum & musique & 2wikimqa & \makecell{multifield\\qa\_en} & \makecell{repobench\\-p} & qasper & hotpotqa & \makecell{multi\_\\news} & trec & \makecell{passage\_\\retrieval\_en} & \makecell{passage\\\_count} & samsum & lcc & Avg.$\uparrow$ \\
\midrule
\textsc{FullAttn} & 33.35 & 86.38 & 18.21 & 19.51 & 19.82 & 33.37 & 49.82 & 58.02 & 41.07 & 43.06 & \textbf{26.57} & 67.00 & 93.50 & 2.97 & 23.15 & 51.86 & 41.73 \\
\midrule
\textsc{MInference} & 32.94 & \textbf{86.87} & 19.46 & 19.57 & 18.85 & 33.30 & 49.14 & 58.98 & \textbf{40.31} & 43.56 & 26.35 & \textbf{68.00} & \textbf{89.00} & 2.10 & 25.58 & 53.68 & 41.73 \\
\textsc{SirLLM} & 32.92 & 85.61 & 21.08 & 21.59 & 24.32 & 34.97 & 48.52 & \textbf{59.15} & 40.17 & 47.00 & 26.44 & 65.50 & 63.00 & 3.00 & 23.11 & 51.83 & 40.51 \\
\textsc{InfLLM} & 25.96 & 84.87 & 20.83 & 19.61 & 13.63 & 27.43 & 41.29 & 55.73 & 30.51 & 38.05 & 25.36 & 64.50 & 10.00 & \textbf{7.50} & 0.28 & \textbf{61.59} & 32.95 \\
\rowcolor{cyan!10}
\textbf{\name} & \textbf{33.46} & 82.39 & \textbf{24.56} & \textbf{23.35} & \textbf{25.12} & \textbf{35.93} & \textbf{52.77} & 57.16 & 40.17 & \textbf{48.70} & 26.41 & 62.00 & 83.00 & 3.00 & \textbf{26.37} & 52.61 & \textbf{42.31} \\
\bottomrule
\end{tabular}
}
\end{table}

\rev{
We also report the maximum memory usage, including the GPU memory, the CPU memory, and the total maximum memory, alongside the average score on LongBench. For \textsc{FullAttn}, we exclude the maximum memory usage, aligning with Figure~\ref{fig:memory_statics}.
}

\begin{table}[H]
\small
\vspace{-5pt}
\caption{\rev{Comparison of methods on LongBench and memory usage.}}
\vspace{5pt}
\label{tab:memory_usage_longbench}
\centering
\scalebox{0.85}{
\begin{tabular}{l|c|c|c|c}
\toprule
Method & LongBench & Max GPU Memory & Max CPU Memory & Total Max Memory \\
\midrule
\textsc{FullAttn} & 41.73 & - & - & - \\
\midrule
\textsc{MInference} & 41.73 & 27.63 & 0.17 & 27.80 \\
\textsc{SirLLM} & 40.51 & 18.29 & \textbf{0.05} & 18.34 \\
\textsc{InfLLM} & 32.95 & 20.03 & 8.95 & 28.98 \\
\rowcolor{cyan!10}
\textbf{\name} & \textbf{42.31} & \textbf{17.71} & 0.15 & \textbf{17.86} \\
\bottomrule
\end{tabular}
}
\end{table}

\rev{
From the experiments above, \name~demonstrates the best overall performance and excels in the majority of subtasks. It outperforms all the baselines without any noticeable performance degradation while consuming less memory. Although MInference also avoids performance drops, it requires more GPU memory compared to \name. SirLLM achieves comparable memory usage but shows some performance decline compared to \textsc{FullAttn} and \name. InfLLM exhibits the most significant performance drop, and its offloading mechanism results in the highest CPU memory usage, making it the method with the largest total memory consumption. These results highlight \name~as an outstanding approach for evaluation on LongBench.
}

\section{\rev{Hyperparameter Analysis}}
\label{sec:hyperanal}
\rev{
We examine three key hyperparameters: budget, stabilizer length, and chunk size.
}

\begin{figure*}[h]
\begin{center}
\subfloat[]{\label{fig:budget}\includegraphics[width=0.3\linewidth]{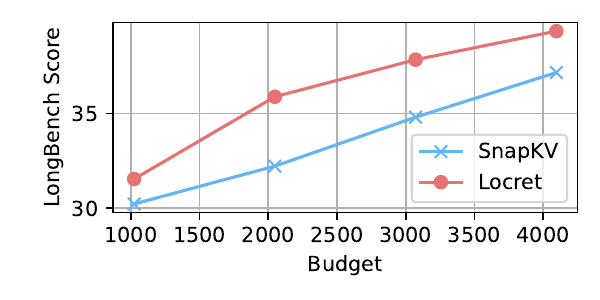}\vspace{-10pt}}
\subfloat[]{\label{fig:stabilizer_nlu}\includegraphics[width=0.3\linewidth]{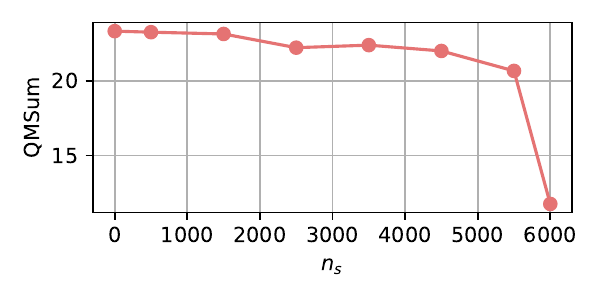}\vspace{-10pt}}
\subfloat[]{\label{fig:chunk_size}\includegraphics[width=0.3\linewidth]{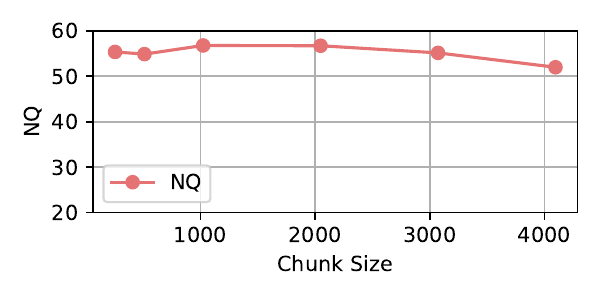}\vspace{-10pt}}
\vspace{-8pt}
\end{center}
\caption{\rev{Scores of \name~under (a) various budgets; (b) various $n_s$; (c) various chunk size.}}
\vspace{-10pt}
\end{figure*}

\rev{
\textbf{Budget} 
To evaluate the robustness of \name~under different budget constraints, we compare the proposed method with \textsc{SnapKV}~\citep{li2024snapkv} using chunked prefill on LongBench~\citep{bai2023longbench}. 
As shown in Figure~\ref{fig:budget},
when the budget size increases, \name~demonstrates a faster performance improvement compared to \textsc{SnapKV}.
}

\textbf{Stabilizers Length} 
As discussed in Figure~\ref{fig:stabilizer}, stabilizers play a crucial role in context retrieval tasks. However, in NLU tasks, the stability of $n_s$ remains relatively high. We evaluate QMSum with different stabilizer lengths $n_s$, with the budget set at 6000. As illustrated in Figure~\ref{fig:stabilizer_nlu}, performance remains consistent when $n_s$ is small. The observed performance degradation at larger $n_s$ values is due to the reduced space available for other cache units.

\textbf{Chunk Size} 
Executing long-context inference on hardware with varying GPU memory limitations choices of chunk size. When the chunk size changes, \name~shows stable performance. We test on the NQ dataset from L-Eval using multiple chunk sizes ranging from 256 to 4096. The results, shown in Figure~\ref{fig:chunk_size}, highlight the stability of $n_s$.

\section{Orthogonality to Other Methods}
\label{sec:ortho}

\label{sec:quant}
\label{sec:pool}

\begin{figure}[H]
    \begin{center}
    \begin{minipage}[t]{0.45\linewidth}  
        \centering  
        \begingroup
            \vspace{-5pt}
            \captionof{table}{Quantization with \textsc{FullAttn} and \name. ``M'' represents Method and ``$-\Delta$'' represents the gap of average L-Eval score.}
            \vspace{5pt}
            \scalebox{0.8}{
            \begin{tabular}{l|cc|c}
            \toprule
            Setting & M & M-4bits & $-\Delta$ \\\midrule
            M=\textsc{FullAttn} & 29.08 & 28.52 & 0.56 \\
            M=\name & 27.96 & 27.11& 0.85 \\\bottomrule
            \end{tabular}
            \label{tab:quant_degradation}
            }
        \endgroup
    \end{minipage}
    \hspace{0.02\linewidth}  
    \begin{minipage}[t]{0.45\linewidth}  
        \centering  
        \begingroup
\vspace{-5pt}
            \captionof{table}{The average L-Eval scores of \textsc{LoCoCo}, \name, and the combination of  \textsc{LoCoCo} and \name. }
\vspace{5pt}
            \scalebox{0.8}{
            \begin{tabular}{l|ccc}
            \toprule
            Method & \textsc{LoCoCo} & \name & \textbf{Combination} \\\midrule
            L-Eval & 26.01 & 27.96 & \textbf{28.70} \\\bottomrule
            \end{tabular}
            \label{tab:lococo}
            }
        \endgroup
    \end{minipage}
    \end{center}
    \vspace{-10pt}
\end{figure}

\textbf{KV cache quantization.} According to~\citet{zhang2024q}, eviction-based methods like \textsc{H$_2$O} struggle with compatibility when combined with KV cache quantization. 
Quantization introduces significant disturbance in the estimation of heavy-hitters, leading to severe performance degradation.
However, \name~is not affected by such issues and can be combined with quantization while maintaining most of its performance. 
Here, we compare the performance degradation caused by quantization on \name~with that of the full attention method using the same metrics. 
We use Quanto as the quantization backend and report the average L-Eval score with \texttt{Llama-3.1-8B-instruct} as the model backbone. 
Table~\ref{tab:quant_degradation} shows that the performance drop caused by quantization on \name~is only slightly higher than that observed with the full attention method, indicating that \name~is a quantization-friendly approach. More details of the experiment are provided in Appendix~\ref{sec:comb_quant}.

\textbf{Token merging.} As described in Section \ref{sec:related_work_compression}, token dropping can also be implemented through an attention pool. 
Attention pool-based methods~\citep{xiao2023efficient, cai2024lococo, mu2024learning, munkhdalai2024leave} merge adjacent tokens or cache units into an attention pool, maintaining a static cache size. 
These methods are orthogonal to \name~, as the evicted tokens can be merged into a small cache pool and retained in GPU memory. 
We conduct the following experiment to demonstrate that \name~can serve as an effective plug-in scoring function within such frameworks, enhancing performance without increasing memory budget.
We select \textsc{LoCoCo}~\citep{cai2024lococo} as a representative of the latest attention pool-based methods. 
\textsc{LoCoCo} maintains a cache set consisting of two parts: the heavy hitters and the convolved non-heavy hitters. 
During each chunked prefill step, \textsc{LoCoCo} first identifies a set of heavy hitters according to \textsc{H$_2$O}~\citep{zhang2024h2o}, then applies 1-D convolution to the non-heavy hitters to compress them into a static size. 
By replacing \textsc{H$_2$O}'s heavy-hitter scoring function with \name, we retain the cache units with high CIS and convolve the others. 
We compare this combination with standalone \textsc{LoCoCo} and \name~on L-Eval using the \texttt{Llama-3.1-8B-instruct} backbone and report the average score across all selected tasks. 
As shown in Table~\ref{tab:lococo}, \name~achieves a higher score than \textsc{LoCoCo}, and the combined algorithm outperforms both standalone methods. 
This suggests that \name~provides a more accurate scoring function compared to \textsc{H$_2$O}, and the two methods complement each other, demonstrating their orthogonality. Further details of the experiment are provided in Appendix~\ref{sec:comb}.

\rev{
\textbf{Head-wise Budget Allocation.} 
Since \name~evict cache units across the attention heads independently, it is compatible with head-wise budget allocation. Here, we combine \name~with \textsc{PyramidKV}~\citep{cai2024pyramidkv}. \textsc{PyramidKV} assumes that identifing the important cache in deeper layers are simpler than shallow layers, thus it allocates more budget to the shallow layers. We evaluate \name+\textsc{PyramidKV} on the following subtasks of $\infty$Bench using \texttt{Phi-3-mini-128K}. Results presented in Figure~\ref{tab:pyramidkv} shows the compatibility of the two methods.
}
\begin{table}[H]
\vspace{-5pt}
\small
\caption{\rev{$\infty$Bench scores of the combination of \name~and \textsc{PyramidKV}.}}
\vspace{5pt}
\label{tab:pyramidkv}
\begin{subtable}{1\textwidth}
\centering
\scalebox{0.85}{
\begin{tabular}{l|cccc|c}
\toprule
\multicolumn{6}{c}{\texttt{Phi-3-mini-128K} on $\infty$Bench} \\
\midrule
Method & R.Number & E.Sum & E.MC & C.Debug & Avg.$\uparrow$\\
\midrule

\name & 97.46 & \textbf{16.82} & 46.29 & 29.71 & 47.57 \\
\name+\textsc{PyramidKV} & \textbf{99.66} & 15.82 & \textbf{48.03} & \textbf{30.00} & \textbf{48.38} \\
\bottomrule

\end{tabular}
}
\end{subtable}

\end{table}

\subsection{Combination with Quantization}
\label{sec:comb_quant}
\begin{table}[H]
\small
\vspace{-5pt}
\caption{L-Eval scores of \textsc{FullAttn}, \textsc{FullAttn}-4bits, \name~and \name-4bits. (Detailed) }
\vspace{5pt}
\label{tab:quant_full}
\begin{subtable}{1\textwidth}
\centering
\scalebox{0.85}{
\begin{tabular}{l|cccccc|c}
\toprule
\multicolumn{8}{c}{\texttt{Llama-3.1-8B-instruct} on L-Eval} \\\midrule
Method & CodeU & NQ & CUAD & NarrativeQA & QMSum & SPACE & Avg.$\uparrow$\\
\midrule

\textsc{FullAttn} & 10.0 & 66.84 & 38.91 & 23.11& 18.76 & 16.86 & 29.08 \\ 
\textsc{FullAttn}-4bits & 7.78 & 66.64 & 38.25 & 22.76 & 18.85 & 16.84 & 28.52 \\ \midrule

\name~ & 8.89 & 63.03 & 37.21 & 23.59 & 18.17 & 16.87 & 27.96\\
\name-4bits & 4.44 & 63.22 & 36.95 & 22.80 & 18.43 & 16.81 & 27.11\\
\bottomrule

\end{tabular}
}
\end{subtable}

\vspace{-10pt}
\end{table}

We compare the combination of \name~and \textsc{HF-4bits} quantization with the full attention method and the standalong \textsc{HF-4bits} quantization. 
We utilize the official implementation of Hugging Face, with Quanto as the backend of quantization. 
Other hyperparameters are kept same as described in Section \ref{sec:setup}. 
We conduct the experiment on L-Eval and report the average score, with \texttt{Llama-3.1-8B-instruct} backend. 
The results in Table~\ref{tab:quant_full} shows that the degradation caused by quantization is not significantly high, showing that \name~exhibits good robustness on data representation and it is friendly to quantization.

\subsection{Combination with \textsc{LoCoCo}}
\label{sec:comb}
\begin{table}[H]
\small
\vspace{-5pt}
\caption{L-Eval scores of \textsc{LoCoCo}, \name~and the combination \textsc{LoCoCo}+\name. (Detailed) }
\vspace{5pt}
\label{tab:comb}
\begin{subtable}{1\textwidth}
\centering
\scalebox{0.85}{
\begin{tabular}{l|cccccc|c}
\toprule
\multicolumn{8}{c}{\texttt{Llama-3.1-8B-instruct} on L-Eval} \\
\midrule
Method & CodeU & NQ & CUAD & NarrativeQA & QMSum & SPACE & Avg.$\uparrow$\\
\midrule

\textsc{FullAttn} & 10.0 & 66.84 & 38.91 & 23.11& 18.76 & 16.86 & 29.08 \\ 
\midrule

\textsc{LoCoCo} & 4.44 & 61.10 & 35.84 & 19.83 & 18.15 & 16.71 & 26.01\\
\name~ & 8.89 & 63.03 & 37.21 & 23.59 & 18.17 & 16.87 & 27.96\\
\textsc{LoCoCo}+\name~ & 7.78 & 66.33 & 38.01 & 24.85 & 18.31 & 16.92 & 28.70\\
\bottomrule

\end{tabular}
}
\end{subtable}

\vspace{-10pt}
\end{table}

We compare the combination of \textsc{LoCoCo} and \name~with the standalone methods.
For \textsc{LoCoCo}, we train the convolution head with the size of convolved cache set to 2048. 
We extend the context length through chunked prefill training to 64K, which is longer than all tasks' average input length.
The convolution kernel is set to 21, and we train the newly-added convolution and layer norms for 200 steps, following the original setting.
Since the original \texttt{Llama-3.1-8B-instruct} supports 128K context length, we do not modify its positional embedding.
During Inference, we keep a cache budget size of 16384. In the standalone \textsc{LoCoCo} setting, there are 2048 cache units are convolved, while the others are the heavy-hitters selected by \textsc{H$_2$O}. 
In the combined algorithm, we replace \textsc{H$_2$O} to \name. 
We select 14336 cache units with the highest CIS, and convolve the other evicted tokens into 2048 cache units. 
In all methods, we set the local length to 0, following the original setting.

\section{Training Robustness}

\begin{figure}[H]
    \begin{center}
    \begin{minipage}[c]{0.45\linewidth}  
        \centering
        \begingroup
        \includegraphics[width=0.8\linewidth]{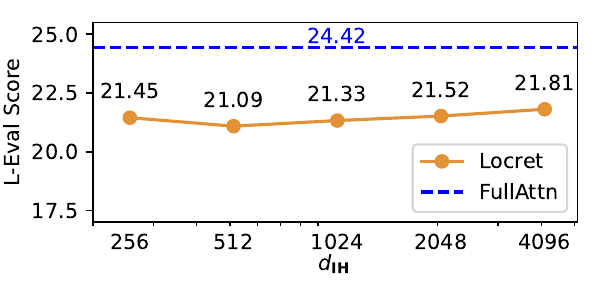}
        \caption{L-Eval scores with different intermediate size of retaining head $d_{\mathbf{R}}$.}
        \label{fig:intermediate}
        \endgroup
    \end{minipage}
    \hspace{0.01\linewidth}  
    \begin{minipage}[c]{0.45\linewidth}  
        \centering  
        \begingroup
\vspace{-5pt}
            \captionof{table}{L-Eval scores of \name~trained on various datasets.}
\vspace{5pt}
            \scalebox{0.75}{
            \begin{tabular}{l|ccc}
            \toprule
            Dataset & LongAlpaca & LongAlign & Anti-Haystack \\
            \midrule
            L-Eval & 21.33 & 22.00 & 20.72 \\
            \bottomrule
            \end{tabular}
            \label{tab:data-insensitivity}
            }
        \endgroup
    \end{minipage}
    \end{center}
\end{figure}

\label{sec:robust}

\name~demonstrates high robustness to the training settings, suggesting that there is no need for careful tuning of training hyperparameters or meticulous selection of datasets. 
Here, we ablate the intermediate size of the retaining heads $d_{\mathbf{R}}$ and train the retaining head $\mathbf{R}$s on various long-context tuning datasets to demonstrate the stability of results across different training settings.

\subsection{Intermediate Size of the retaining head}

We align all the training settings as described in Section~\ref{sec:setup} and only change the intermediate size of retaining heads $d_{\mathbf{R}}\in \{256, 512, 1024, 2048, 4096\}$ with the backbone model \texttt{Phi-3-mini-128K}. 
The trained model is evaluated on L-Eval and we report the average L-Eval score corresponding to each intermediate size. 
Results are listed in Figure~\ref{fig:intermediate}. 
The performance variations among all the settings are minimal compared to the changes in the intermediate size, surpassing all baselines in Table~\ref{tab:results}.
This indicates that out method exhibits good performance stability regardless of the intermediate size of the retaining head $\mathbf{R}$s.

\begin{table}[H]
\small
\vspace{-5pt}
\caption{L-Eval scores with different intermediate size of the retaining head $d_{\mathbf{R}}$. (Detailed)}
\vspace{5pt}
\label{tab:intermediate}
\begin{subtable}{1\textwidth}
\centering
\scalebox{0.85}{
\begin{tabular}{l|cccccc|c}
\toprule
\multicolumn{8}{c}{\texttt{Phi-3-mini-128K} on L-Eval} \\
\midrule
$d_{\mathbf{R}}$ & CodeU & NQ & CUAD & NarrativeQA & QMSum & SPACE & Avg.$\uparrow$\\
\midrule

256 & 8.89 & 51.52 & 23.05 & 16.21 & 15.26 & 13.77 & 21.45\\
512 & 6.67 & 50.61 & 23.33 & 16.67 & 15.02 & 14.23 & 21.09\\
1024 & 8.89 & 51.49 & 22.23 & 16.42 & 14.86 & 14.06 & 21.33\\
2048 & 7.78 & 54.09 & 21.91 & 16.46 & 15.00 & 13.89 & 21.52\\
4096 & 10.00 & 52.33 & 23.52 & 16.15 & 14.81 & 14.02 & 21.81\\
\bottomrule
\end{tabular}
}
\end{subtable}

\vspace{-10pt}
\end{table}

We train different retaining head $\mathbf{R}$s with $d_{\mathbf{R}}\in\{256, 512, 1024, 2048, 4096\}$. 
We keep all the other hyperparameters same, and train on the same dataset.
From Table \ref{tab:intermediate}, \name~shows stability to the intermediate size, in both overall performance and the performance of each single task.
While increasing the intermediate size, we observe very slight overall performance enhancement.
However, the performance variance is negligible compared to the increase of parameter size, thus we choose to maintain the intermediate size in a small scope to take balance of performance and efficiency.

\subsection{Training Data Insensitivity}

We also consider the sensitivity of the training data, which leads us to ablate the training dataset by training on LongAlign~\citep{bai2024longalign} and Anti-Haystack~\citep{anti-haystack}, comparing these results with those from LongAlpaca~\citep{chen2023longlora} in the original training setting. 
We also align other settings to the original setting and choose the backbone model to be \texttt{Phi-3-mini-128K}. 
We report the average L-Eval score for each training dataset. 
The results in Table~\ref{tab:data-insensitivity} shows that \name~has high insensitivity towards different training data. 
The performance impact of different data recipes is minimal, indicating that our method can be trained on any long-context tuning dataset.

\begin{table}[H]
\small
\vspace{-5pt}
\caption{L-Eval scores of \name~trained on various dataset. (Detailed)}
\vspace{5pt}
\label{tab:datasets}
\begin{subtable}{1\textwidth}
\centering
\scalebox{0.85}{
\begin{tabular}{l|cccccc|c}
\toprule
\multicolumn{8}{c}{\texttt{Phi-3-mini-128K} on L-Eval} \\
\midrule
Dataset & CodeU & NQ & CUAD & NarrativeQA & QMSum & SPACE & Avg.$\uparrow$\\
\midrule
LongAlpaca & 8.89 & 51.49 & 22.23 & 16.42 & 14.86 & 14.06 & 21.33\\
LongAlign & 10.00 & 55.13 & 21.34 & 16.40 & 15.01 & 14.09 & 22.00\\
Anti-Haystack & 8.89 & 52.91 & 20.87 & 13.73 & 13.84 & 14.10 & 20.72\\
\bottomrule
\end{tabular}
}
\end{subtable}
\vspace{-10pt}
\end{table}

We conduct training on various datasets and benchmark the trained weights on L-Eval with \texttt{Phi-3-mini-128K} backbone, to show the stability towards training datasets. 
For each datasets, we set the training hyperparameters same and truncate the context to 10240 tokens. 
We train the first 3000 steps of LongAlpaca and LongAlign. 
Since Anti-Haystack is a relatively smaller dataset, we utilize the whole dataset, which consist of 2424 entries. 
The results in Table \ref{tab:datasets} shows that different training dataset recipe exhibits minor effect towards the overall performance. 
\name~can obtain competitive performance without delicately selecting the training data.

\section{\rev{Extremely Long Context Evaluation}}
\label{append:extreme}

\rev{
We create a dataset similar to $\infty$Bench's R.Number, with an average length of 10 million tokens. Each data point contains a 10-digit number string inserted into an irrelevant context, and the task is to retrieve the inserted number. The dataset consists of 50 examples, with the number strings uniformly distributed throughout the context. We used the hyperparameters from Table~\ref{tab:inf-hyper}, with the exception of setting the chunk size to 10240 to speed up inference. The results, presented below in Table~\ref{tab:extreme}, show that Locret can efficiently process extremely long contexts. In this experiment, the cache budget is set to 6000, and the compression ratio is 1747.6$\times$.
}

\begin{table}[H]
\small
\vspace{-5pt}
\caption{\rev{Inference speed with Retaining Heads.}}
\vspace{5pt}
\label{tab:extreme}
\begin{subtable}{1\textwidth}
\centering
\scalebox{0.9}{
\begin{tabular}{l|c}
\toprule
\multicolumn{2}{c}{\texttt{Phi-3-mini-128K} on 10M context} \\
\midrule
Dataset & R.PassKey\_10M \\
\midrule
\name & 100.00 \\
\bottomrule
\end{tabular}
}
\end{subtable}
\vspace{-10pt}
\end{table}

\section{\rev{Compressing Multi-turn Conversations}}

\label{append:multi-turn}
\rev{
Compared to query-aware eviction methods, such as \textsc{SnapKV}~\citep{li2024snapkv}, \name~is a more suitable solution for multi-turn conversation scenarios. This is because the evaluation of cache importance in \name~is based on the cache itself, rather than being dependent on the subsequent query. To demonstrate this, we evaluate \name~on the Rock-Paper-Scissors benchmark introduced in \textsc{SirLLM}~\citep{yao2024sirllm}. Since \textsc{SirLLM} is specifically designed for such scenarios, we use it as our baseline in this benchmark. Results in Table~\ref{tab:rps} show that Locret is also effective in multi-turn conversation contexts. 
}

\rev{
The hyperparameters are aligned with those used in \textsc{SirLLM}, with the cache budget set to 1024, and no stabilizers are retained, as \textsc{SirLLM} does not retain local tokens in this benchmark. We perform 2000 turns as same as the original \textsc{SirLLM} settings. The results are presented below.
}

\begin{table}[H]
\small
\vspace{-5pt}
\caption{\rev{Rock-Paper-Scissors scores of \name~and \textsc{SirLLM}.}}
\vspace{5pt}
\label{tab:rps}
\begin{subtable}{1\textwidth}
\centering
\scalebox{0.9}{
\begin{tabular}{l|ccccccccc|cc}
\toprule
\multicolumn{12}{c}{\texttt{Phi-3-mini-128K} on Rock-Paper-Scissors} \\
\midrule

\multirow{2}{*}{Preference} & \multicolumn{3}{c}{Rock} & \multicolumn{3}{c}{Paper} & \multicolumn{3}{c|}{Scissors} & \multicolumn{2}{c}{Avg.} \\
 & win & tie & lose & win & tie & lose & win & tie & lose & win$\uparrow$ & lose$\downarrow$ \\
\midrule
\textsc{SirLLM} & 40.00 & 31.75 & 28.25 & 27.5 & 36.55 & 35.96 & 29.35 & 25.15 & 45.50 & 32.28 & 36.57 \\
\rowcolor{cyan!10}
\textbf{\name} & 18.95 & 50.00 & 31.05 & 30.35 & 19.45 & 50.20 & 52.05 & 27.25 & 20.70 & \textbf{33.78} & \textbf{33.98} \\

\bottomrule
\end{tabular}
}
\end{subtable}
\end{table}

\section{Discontinuous Context and Stablizers}
\label{sec:discontinuous-context}

Evicting cache units results in context discontinuity, which causes unstable CIS prediction and inaccurate calculation of later tokens. 
Thus, we always retain the stabilizers, which are consist of the last $n_{s}$ cache units in each chunked prefill step. 
We ablate $n_{s}$ on R.Number of $\infty$-Bench in the proposed algorithm to demonstrate the necessity of incorporating stabilizers in the design. 
The results in Figure~\ref{fig:stabilizer_a} show that lower stabilizer length $n_s$ causes severe performance degredation and the model fails completely when the stabilizers are absent. 
We report the maximum absolute error of the last hidden state of the input prompt across different layers in Figure~\ref{fig:stabilizer_b}. 
Large errors can be observed when the stabilizers are short or absent. 
We also report the mean absolute error of the predicted causal importance values with different stabilizer lengths, compared to the case without evicting any cache units, in Figure~\ref{fig:stabilizer_c}. 
We also observe high errors when the stabilizer length is limited. 
This explains the reason for failure when the stabilizers are short or absent: context discontinuity leads to instability in the prediction of CIS, resulting in errors during cache eviction and amplifying errors in the hidden states.

\section{\rev{Retaining Heads Do not Slow Down Inference}}
\label{append:time}
\rev{
We evaluate the model's forward throughput under varying context lengths, both with and without retaining heads. The results are summarized below in Table~\ref{tab:overhead}. ``$\mathbf{R}$'' represents the retaining heads, and the throughput is reported in tokens per second (tok/s) in the format ``Avg. / Std.''}

\begin{table}[H]
\small
\vspace{-5pt}
\caption{\rev{Inference speed with Retaining Heads.}}
\vspace{5pt}
\label{tab:overhead}
\begin{subtable}{1\textwidth}
\centering
\scalebox{0.9}{
\begin{tabular}{l|ccccc}
\toprule
Context Length & 1024 & 2048 & 3072 & 4096 \\
\midrule
w/o $\mathbf{R}$ Speed & 18674 / 443 & 19743 / 464 & 19982 / 402 & 20304 / 187 \\
w/ $\mathbf{R}$ Speed & 17118 / 1117 & 18503 / 546 & 19054 / 283 & 19153 / 174\\
\bottomrule
\end{tabular}
}
\end{subtable}
\vspace{-10pt}
\end{table}
\rev{
From the results, no significant latency increase is observed when using retaining heads. The numerical differences are attributed to systematic variations rather than additional overhead introduced by retaining heads during inference.}

\section{Causal Importance Score Simulates a Cache Problem}
\label{append:causal_importance}

In this section, we show that assigning each cache unit a CIS and calculate each cache units with top-$b$ cache units simulates a cache problem, i.e. the calculation process can be done in a cache. Thus, \name~mathmatically equals to top-$b$ sparse attention.

\begin{definition}{(Causal Calculation)}
    Given a sequence of objects $c_1, c_2, \cdots, c_n$, if 
    \begin{center}
    $\forall 1\leq i\leq n$, $c_i = f(c_1, c_2, \cdots, c_{i-1})$
    \end{center}
    then $f$ is a causal calculation. $c_1, c_2, \cdots, c_{n}$ is the generated sequence respective to $f$.
\end{definition}

For all causal calculations, we can easily split the function into two parts: a selection function and a another function. Formally, 
\begin{center}
    $\forall \text{ causal calculation } $f$,~\exists \text{ function }g, Sel,$ 

    $g: 2^{\{c_1, c_2, \cdots, c_n\}}\rightarrow\{c_1, c_2, \cdots, c_n\},$

    $Sel: 2^{\{c_1, c_2, \cdots, c_n\}}\rightarrow2^{\{c_1, c_2, \cdots, c_n\}};~ X\mapsto Y \subseteq X,$

    $\text{s.t. } f = g\circ Sel.$
\end{center}

\begin{definition}{(Causal Importance Score)}
    Given a causal calculation $f$ and $c_1, c_2, \cdots, c_n$ is the generated sequence of $f$. $s_1, s_2, \cdots, s_n\in \mathbb{R}$ is a sequence of numbers. If 
    \begin{center}
        $s_i = h(c_i),$
    \end{center}
    then $\{s_i\}$ is a CIS of sequence $\{c_i\}$. $h$ is a causal importance scoring function. 
\end{definition}

\begin{definition}{(Cache Problem)}
    Given a causal calculation $f = g\circ Sel$, its generated sequence $\{c_i\}$ and a positive number $b\in \mathbb{Z}_+$, if $f$ satisfies the following two condion, then $(f, b, \{c_i\})$ is a cache problem with budget $b$.
    \begin{itemize}
        \item $\forall 1\leq i\leq n,~ |Sel(c_1, \cdots, c_n)|\leq b,$
        \item $\forall 1\leq m_1 < m_2 \leq n,~ Sel(c_1, \cdots, c_{m_2}) \backslash Sel(c_1, \cdots, c_{m_1})\subseteq\{c_{m_1+1}, \cdots, c_{m_2}\}.$
    \end{itemize}
\end{definition}

\begin{theorem}{(Calculating cache units with Top-$b$ CIS is a cache problem.)}
    Given a causal calculation $f = g\circ Sel$, and its generated sequence $\{c_i\}$, a CIS $s_i = h(c_i)$ and a positive number $b\in \mathbb{Z}_+$, if the selection function $Sel$ satisfies the following condition,
    \begin{center}
        $Sel(c_1, c_2, \cdots, c_i)=\{c_{p_1}, c_{p_2} \cdots, c_{p_{b'}}\},~s_{p_1}, s_{p_2} \cdots, s_{p_{b'}}
        \in \text{Top-}b(s_1, s_2, \cdots, s_i)$
    \end{center}
    then $(f, b, \{c_i\})$ is a cache problem with budget $b$.
\end{theorem}

\textbf{Proof.} 
(1) For all $i$ of $1\leq i \leq n$, $|Sel(c_1, \cdots, c_i)| =|\{c_{p_1} \cdots, c_{p_{b'}}\}| = |\{s_{p_1}, \cdots, s_{p_{b'}}\}|$. Since $s_{p_1}, s_{p_2} \cdots, s_{p_{b'}}\in \text{Top-}b(s_1, s_2, \cdots, s_i)$, $|\{s_{p_1}, \cdots, s_{p_{b'}}\}|\leq b$. Thus $|Sel(c_1, \cdots, c_i)|\leq b$.

(2) For all $1 \leq m_1, < m_2, \leq n$, 
\begin{align*}
    Sel(c_1, \cdots, c_{m_2}) \backslash Sel(c_1, \cdots, c_{m_1})&\subseteq\{c_{m_1+1}, \cdots, c_{m_2}\}\\\iff \{s_{p_1}, \cdots, s_{p_{m_2}}\}\backslash \{s_{q_1}, \cdots, s_{q_{m_1}}\}&\subseteq \{s_{m_1+1}, \cdots, s_{m_2}\}.
\end{align*}
Assume $\exists s\in \{s_{p_1}, \cdots, s_{p_{m_2}}\}\backslash \{s_{q_1}, \cdots, s_{q_{m_1}}\}$ but $s\notin \{s_{m_1+1}, \cdots, s_{m_2}\}$. Since $s_{p_1}, \cdots, s_{p_{m_2}} = \text{Top-}b(s_1, \cdots, s_{m_2}), ~s\in \{s_1, \cdots, s_{m_2}\}$. Thus $s\in \{s_1, \cdots, s_{m_1}\}$. $s$ is not in the Top-$b$ values of first $m_1$ scores, thus there exists $b$ values larger than $s$, denote as $s_{l_1}, \cdots, s_{l_b}$. Then, $s_{p_1}, \cdots, s_{p_{m_2}} = \text{Top-}b(s_{l_1}, \cdots, s_{l_b}, s_{m_1+1}, \cdots, s_{m_2}).$ From this, we can obtain that $\min\{s_{p_{m_2}}\}\geq \min\{s_{l_1}, \cdots, s_{l_b}\} > s$, $s\notin \{s_{p_1}, \cdots, s_{p_{m_2}}\}$. Contradiction.  Finally, there must be $s\in \{s_{m_1+1}, \cdots, s_{m_2}\}$.
From (1)(2), $f$ satisfies the two conditions of cache problem. Thus, calculating cache units with Top-$b$ CIS is a cache problem.

\section{Retained Patterns of \name~}
\label{sec:retained_patterns}

\begin{figure}[h]
    \vspace{-10pt}
    \begin{center}
    \begin{minipage}[t]{\linewidth}  
        \centering
        \begingroup
        \includegraphics[width=0.8\linewidth]{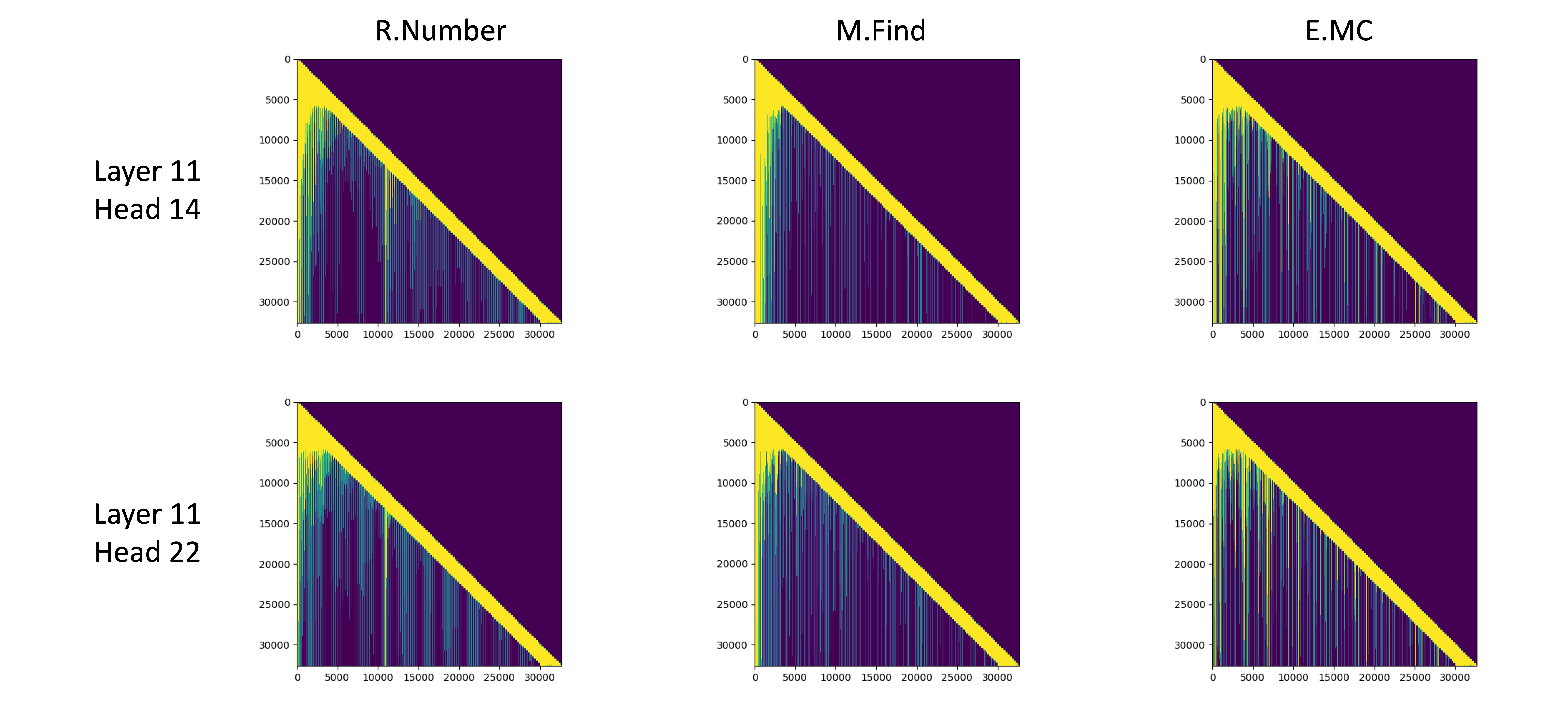}
        \caption{Head patterns across multiple tasks.}
        \label{fig:head_pattern}
        \endgroup
    \end{minipage}
    \begin{minipage}[t]{\linewidth}  
        \centering
        \begingroup
        \includegraphics[width=0.9\linewidth]{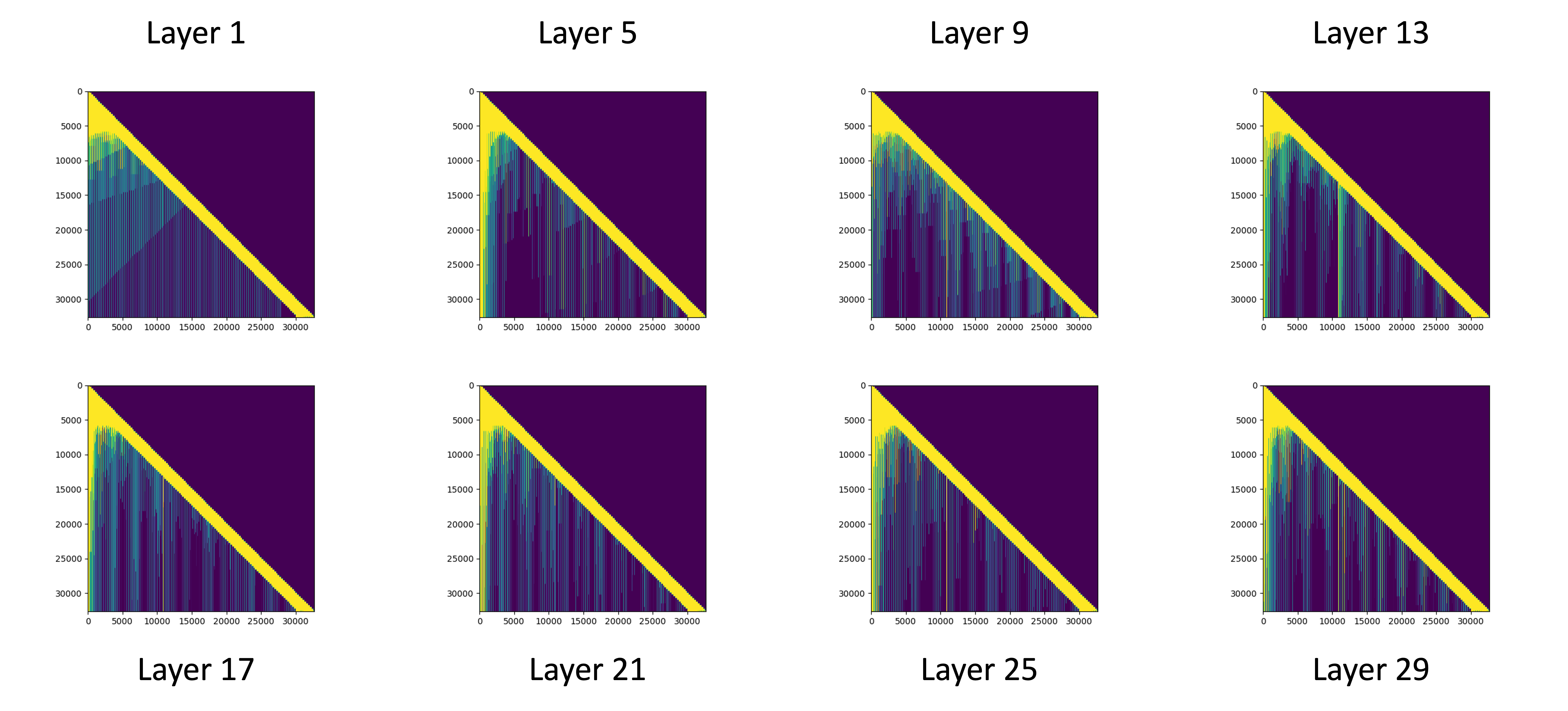}
        \caption{Layer patterns of R.Number}
        \label{fig:layer_pattern}
        \endgroup
    \end{minipage}
    \end{center}
\end{figure}

We investigate the retained patterns of \name. 
We trace the cache units at each attention head through the chunked prefill on R.Number, M.find and E.MC of $\infty$Bench with backbone \texttt{Phi-3-mini-128K}, and investigate the pattern variation among different layers on R.Number.
We display the results in Figure~\ref{fig:head_pattern} and Figure~\ref{fig:layer_pattern}. 
The yellow parts are the retained cache, where the y-axis represents cache position and x-axis is the time axis. 

Figure \ref{fig:head_pattern} shows that the pattern is mostly decided by the tasks, where both heads shows similar pattern in the same task. 
In R.Number, we are able to observe a strong signal between token 10000 and 15000, which is the position of the inserted number string, indicating that \name~can identify the potentially answer-related parts by giving high predicted values of CIS.
In M.Find, we can observe the StreamingLLM~\citep{xiao2023efficient} pattern, where the tokens at the beginning of the sequence are always important. 
This is also mentioned as the $\Lambda$-pattern in \textsc{MInference}.
We can also discover the vertical lines in the middle of the sequence. 
This pattern is also approached by \textsc{MInference}~\citep{jiang2024minference} by the pattern ``vertical-and-slash''.
In E.MC, \textsc{H$_2$O}~\citep{zhang2024h2o} and ScissorHands~\citep{liu2024scissorhands} pattern can be observed, following the assumption that if a token is activated at some point, it will continue to be activated in the consequencing process. 
Noticing that the vertical lines always come in groups, which is the fundament of \textsc{InfLLM}~\citep{xiao2024infllm} retrieving blocks to calculate. The comparison between two heads also shows that different heads exhibits different features. 
Head 22 of layer 11 shows stronger vertical lines at some point, where retained pattern of head 14 layer 11 is more even. 
Head 14 of layer 11 also gives stronger signal to the initial tokens, where this effect is less strong in head 22 layer 11. 
We also conduct experiments to investigate the patterns across layers. 
In Figure~\ref{fig:layer_pattern}, we show that the pattern variance of the same head in different layers can be large. 
In shallow layers, e.g. layer 1 and 5, the retained cache units appears to be periodical and semantic independent.
However, in middle layers, e.g. layer 13 and 17, the position of the inserted number string is strongly highlighted, indicating that semantic takes over to be the dominant factor. 
In the deepest layers, e.g. 21, 25 and 29, the highlighted vertical line at the position of the inserted string becomes more accurate. 

The retained pattern at different layers shows various features, which might be a good handle to investigate how LLMs understand and process natural language queries. 

\section{The Loss Dynamics of \name}
\label{sec:training_dynamics}
\begin{figure}[h]
\vspace{-10pt}
\begin{center}
\includegraphics[width=0.9\linewidth]{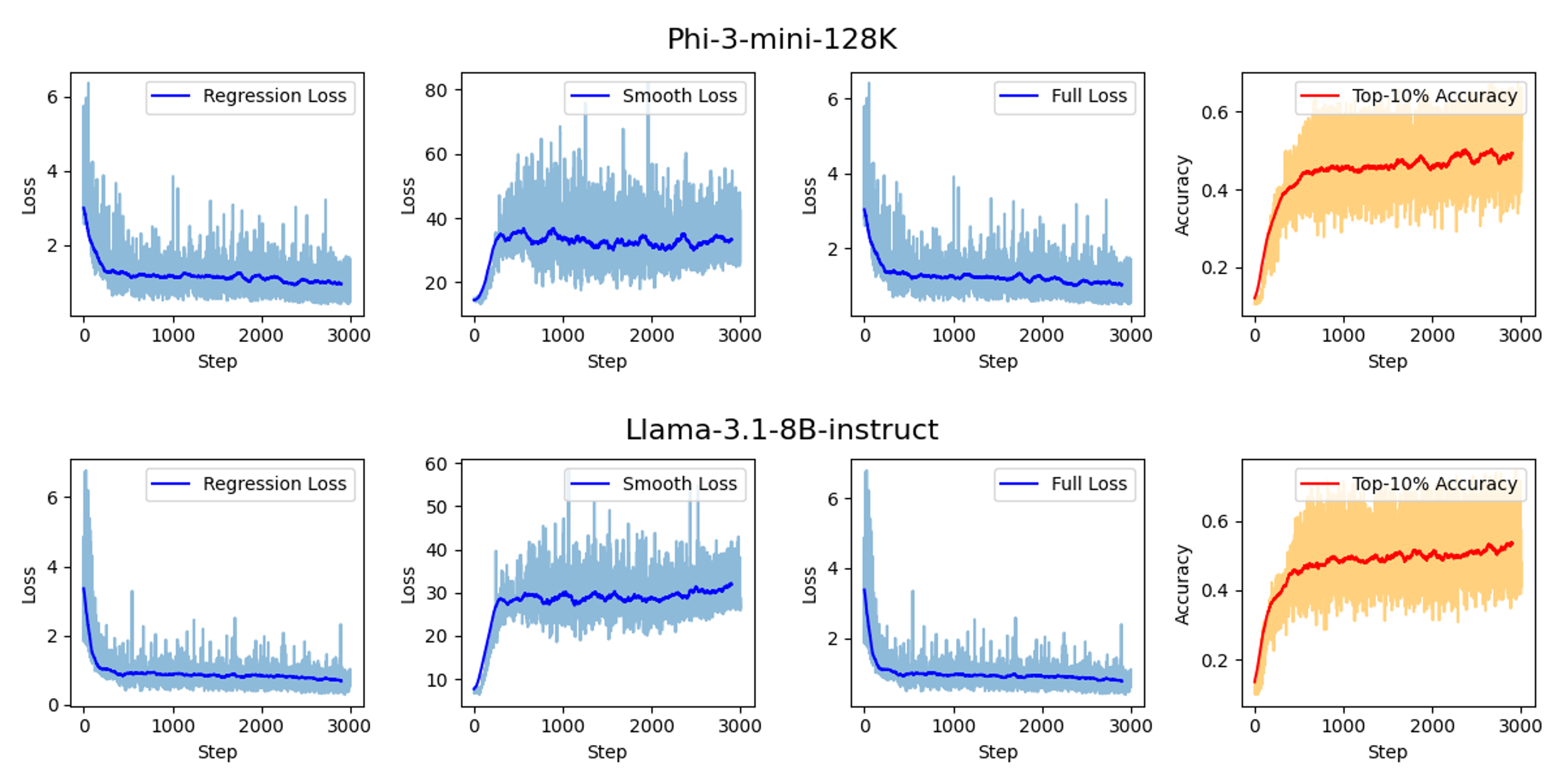}
\end{center}
\caption{Training loss and accuracy during the training process.}
\label{fig:training_dynamics}
\end{figure}

Here, we present changing trend of loss and accuracy during training in Figure~\ref{fig:training_dynamics}.

\end{document}